\documentclass{neus2025}

\usepackage{bbm}
\usepackage{multirow}
\usepackage{booktabs}
\usepackage{makecell}
\usepackage{pythonhighlight}
\usepackage{enumitem}
\usepackage{adjustbox}
\usepackage{pdflscape}
\usepackage{hyperref}

\title[SpecRLBench]{SpecRLBench: A Benchmark for Generalization in Specification-Guided Reinforcement Learning}
\usepackage{times}

\author{%
 \Name{Zijian Guo} \Email{zjguo@bu.edu}\\
 \addr Boston University
 \AND
 \Name{\.Ilker I\c{s}{\i}k} \Email{iilker@bu.edu } \\
 \addr Boston University
 \AND
 \Name{H. M. Sabbir Ahmad} \Email{sabbir92@bu.edu} \\
 \addr Boston University
 \AND
 \Name{Wenchao Li} \Email{wenchao@bu.edu}\\
 \addr Boston University
}

\newcommand{\always}{\mathsf{G}\,}
\newcommand{\event}{\mathsf{F}\,}
\newcommand{\until}{\;\mathsf{U}\;}

\begin{document}

\maketitle

\vspace{-5mm}
\begin{abstract}
  Specification-guided reinforcement learning (RL) provides a principled framework for encoding complex, temporally extended tasks using formal specifications such as linear temporal logic (LTL). 
  While recent methods have shown promising results, their ability to generalize across unseen specifications and diverse environments remains insufficiently understood. 
  In this work, we introduce SpecRLBench, a benchmark designed to evaluate the generalization capabilities of LTL-based specification-guided RL methods. 
  The benchmark spans multiple difficulty levels across navigation and manipulation domains, incorporating both static and dynamic environments, diverse robot dynamics, and varied observation modalities. 
  Through extensive empirical evaluation, we characterize the strengths and limitations of existing approaches and reveal the challenges that emerge as specification and environment complexity increase. 
  SpecRLBench provides a structured platform for systematic comparison and supports the development of more generalizable specification-guided RL methods.
  Code is available at \href{https://github.com/BU-DEPEND-Lab/SpecRLBench}{https://github.com/BU-DEPEND-Lab/SpecRLBench}
  
\end{abstract}


\section{Introduction}
\label{sec:intro}


Reinforcement learning (RL) enables agents to learn decision-making policies through interaction with their environments.
Numerous studies have achieved promising results in domains such as robotics~\citep{jiang2024learning, luo2025precise}, autonomous driving~\citep{wang2023vision, zhang2025carplanner, lu2025preference}, and healthcare~\citep{gorrepati2025reinforcement, choppara2025efficient}. 
However, effectively handling complex tasks with long-term temporal structure remains a significant challenge for most existing approaches.
For example, a manipulation task may require an arm to reach multiple target regions in a specific order while always avoiding unsafe areas.

\vspace{2mm}
\noindent To specify such complex tasks, various approaches have been proposed. One prominent direction uses natural language instructions~\citep{yang2021safe, mees2022calvin, huang2025vlm, zakharov2025goalladder, kim2025fine, yu2025rlinf}, which provide a flexible and intuitive interface for describing desired behaviors. 
While effective in many settings, natural language specifications can be ambiguous and may lack precise semantics~\citep{pang2023natural, kharyal2024glide, xie2025dail}, making consistent interpretation and verification challenging.

\vspace{2mm}
\noindent
An alternative line of work employs formal specification languages~\citep{araki2021logical, voloshin2023eventual, jothimurugan25a}, which offer precise, unambiguous descriptions of desired behaviors and explicitly capture temporal structure.
While specification-guided RL has attracted growing interest, progress in this area is hindered by the lack of a standard benchmark.
Existing methods are typically evaluated in isolated environments with limited task diversity, making it difficult to systematically compare approaches or to assess key capabilities such as generalization to unseen specifications~\citep{jackermeier2025deepltl, guo2025one} and robustness to environment variations~\citep{fan2025imitation, bagatella2025directed}.

\vspace{2mm}
\noindent 
To address this gap, we introduce our benchmark, the \textbf{Specification-Guided RL Benchmark (SpecRLBench)}, which provides a diverse set of environments for training and testing specification-guided RL methods, with the goal of enabling systematic evaluation of generalization.
Overall, SpecRLBench consists of 19 environment variants spanning multiple difficulty levels across navigation and manipulation domains, incorporating static and dynamic environments, diverse robot dynamics, and varied observation modalities.
All environments follow the standard Gym interface~\citep{towers2024gymnasium}, allowing seamless integration with existing methods and serving as a convenient starting point for developing and testing new ideas.
In addition, we conduct a comprehensive empirical study and provide insights into the strengths and limitations of current approaches.



\section{Related Work}
\label{sec:related-work}




\noindent
\textbf{Specification-guided RL.} A growing body of work studies RL under explicit task specifications, where desired behaviors are defined using structured or formal representations.
Among these, temporal logic has been widely adopted to encode long-horizon objectives and safety-critical requirements, including linear temporal logic (LTL)~\citep{cai2023overcoming, tasse2024skill, manganaris2025automaton, fan2025imitation, shah2025ltlconstrained}, 
signal temporal logic (STL)~\citep{aksaray2016q, guo2024temporal, xiong2024co, wang2024tractable, meng2025telograf, meng2025tgpo, liu2025learning}, 
and other related formalisms~\citep{li2017reinforcement, brafman2018ltlf, jothimurugan2019composable, de2019foundations, jothimurugan2021compositional, furelos2023hierarchies, shukla2024logical, roy2025learning}.
Temporal logic has also been used to specify coordinated behaviors and interaction constraints in multi-agent settings to guide policy learning~\citep{eappen2022distspectrl, hammond2021multi, wang2023multi, ardon2023learning, smith2023automatic, yalcinkaya2025automata}.
Despite these advances, most existing methods focus on learning policies for a single fixed specification~\citep{hasanbeig2018logically, hahn2019omega, bozkurt2020control, icarte2022reward, zhou2022hierarchical, shao2023sample, le2024reinforcement}.
More recently, the expressiveness of temporal logic and the need to adapt to changing task requirements in real-world applications have drawn increasing attention to generalization across specifications~\citep{vaezipoor2021ltl2action, qiu2023instructing, yalcinkaya2024compositional, jackermeier2025deepltl, guo2025one, meng2025telograf}.
Collectively, these methods highlight the potential of formal specifications for capturing complex behaviors. 
However, existing evaluations are often conducted within a single environment or a narrowly varying set of environments, limiting insight into how specification-guided methods generalize and scale beyond a particular setting.

\vspace{2mm}
\noindent
\textbf{Existing benchmarks.}
The study of generalization in specification-guided RL is closely related to several areas, including multi-task RL~\citep{vithayathil2020survey}, as each specification can be viewed as defining a task, and goal-conditioned RL~\citep{liu2022goal}, since the representation of a specification effectively serves as the goal signal for policy learning.
A number of benchmarks have been proposed in these domains, such as Meta-World~\citep{yu2020meta}, CompoSuite~\citep{mendez2022composuite}, and CORA~\citep{powers2022cora} for multi-task RL, and MiniGrid~\citep{MinigridMiniworld23}, BabyAI~\citep{chevalier2018babyai}, D4RL~\citep{fu2020d4rl}, and OGBench~\citep{park2024ogbench} for goal-conditioned RL.
However, benchmarks such as Meta-World, CompoSuite, and CORA do not explicitly focus on tasks with temporal and logical constraints. Their tasks are typically defined as single-step objectives, such as picking an object or opening a door, and rarely require sequential or temporally structured behaviors.
Similarly, D4RL and OGBench primarily emphasize goal-reaching without incorporating temporal and logic requirements.
While MiniGrid and BabyAI include instruction-following tasks with some sequential structure, they are limited to discrete-action environments with grid-based observations, and do not cover high-dimensional, continuous-control settings.
There also exist benchmarks based on natural language conditioning, such as CALVIN~\citep{mees2022calvin}, LIBERO~\citep{liu2023libero}, and ImagineBench~\citep{pang2025imaginebench}. However, these environments are typically designed to evaluate advanced vision-language or vision-language-action models, and thus diverge from the current focus on studying generalization in specification-guided RL under formal specifications.


\vspace{-2mm}
\section{Preliminaries}
\label{sec:preliminaries}

\textbf{Reinforcement Learning.} We model the decision-making problem as a Markov decision process (MDP) defined by the tuple
$\mathcal{M} := (\mathcal{S}, \mathcal{A}, P, r, \gamma, d_0),$
where $\mathcal{S}$ denotes the state space, $\mathcal{A}$ is the action space,
$P: \mathcal{S} \times \mathcal{A} \times \mathcal{S} \rightarrow [0,1]$ is the transition dynamics,
$r: \mathcal{S} \times \mathcal{A} \rightarrow \mathbb{R}$ is the reward function,
$\gamma \in (0,1)$ is the discount factor, and
$d_0 \in \Delta(\mathcal{S})$ denotes the initial state distribution.
Let $\pi: \mathcal{S} \times \mathcal{A} \mapsto [0, 1]$ denote the policy and $\tau = \{ s_t, a_t, r_t \}_{t=0}^\infty$ denote the trajectory through interaction with the environment, where $r_t = r(s_t, a_t)$.
In standard RL, the goal is to find the optimal policy $\pi^*$ that maximizes the expected discounted cumulative reward: $\max_{\pi} \mathbb{E}_{\tau \sim \pi}[\sum_{t=0}^\infty \gamma^tr_t]$.

\vspace{2mm}
\noindent \textbf{Linear temporal logic.} Linear Temporal Logic (LTL)~\citep{pnueli1977temporal} provides a formal framework for specifying temporal properties over infinite sequences of system states.
An LTL formula is constructed from a finite set of atomic propositions $AP$ using Boolean operators, including negation ($\neg$), conjunction ($\land$), and disjunction ($\lor$), together with temporal operators such as "until" ($\mathsf{U}$), "eventually" ($\mathsf{F}$), and "always" ($\mathsf{G}$).
Formally, for $\boldsymbol{a} \in AP$, the syntax is defined recursively as

\vspace{-3mm}
\begin{equation*}
    \varphi := \boldsymbol{a} \mid \neg \varphi \mid \varphi_1 \land \varphi_2 \mid \varphi_1 \lor \varphi_2 \mid \event \varphi \mid \always \varphi \mid \varphi_1 \until \varphi_2 .
\end{equation*}

\noindent
Intuitively, the formula $\varphi_1 \until \varphi_2$ is satisfied if $\varphi_2$ holds at some future time step and $\varphi_1$ holds at all preceding steps.
The temporal operator $\event \varphi$ requires that $\varphi$ be satisfied at least once in the future, whereas $\always \varphi$ requires $\varphi$ to hold at all time steps from the current point onward. To interpret LTL formulas in an MDP, a labeling function is typically assumed $L: \mathcal{S} \rightarrow 2^{AP}$, which associates each state with the set of atomic propositions that are true in that state.
A trajectory $\tau$ thereby induces a trace $\mathrm{Tr}(\tau) = L(s_0)L(s_1)\ldots$, and we write $\tau \models \varphi$ if this trace satisfies the LTL formula $\varphi$.
Given a policy $\pi$, the probability of satisfying $\varphi$ is defined as
$\Pr(\pi \models \varphi)
= \mathbb{E}_{\tau \sim \pi}\!\left[ \mathbbm{1}[\tau \models \varphi] \right],$
where $\mathbbm{1}[\cdot]$ denotes the indicator function.
In this context, an optimal policy is one that maximizes the probability of satisfying the given LTL specification.

\vspace{-2mm}
\section{Benchmark}
\label{sec:benchmark}

\begin{figure}[ht]
    \centering
    \includegraphics[width=0.9\linewidth]{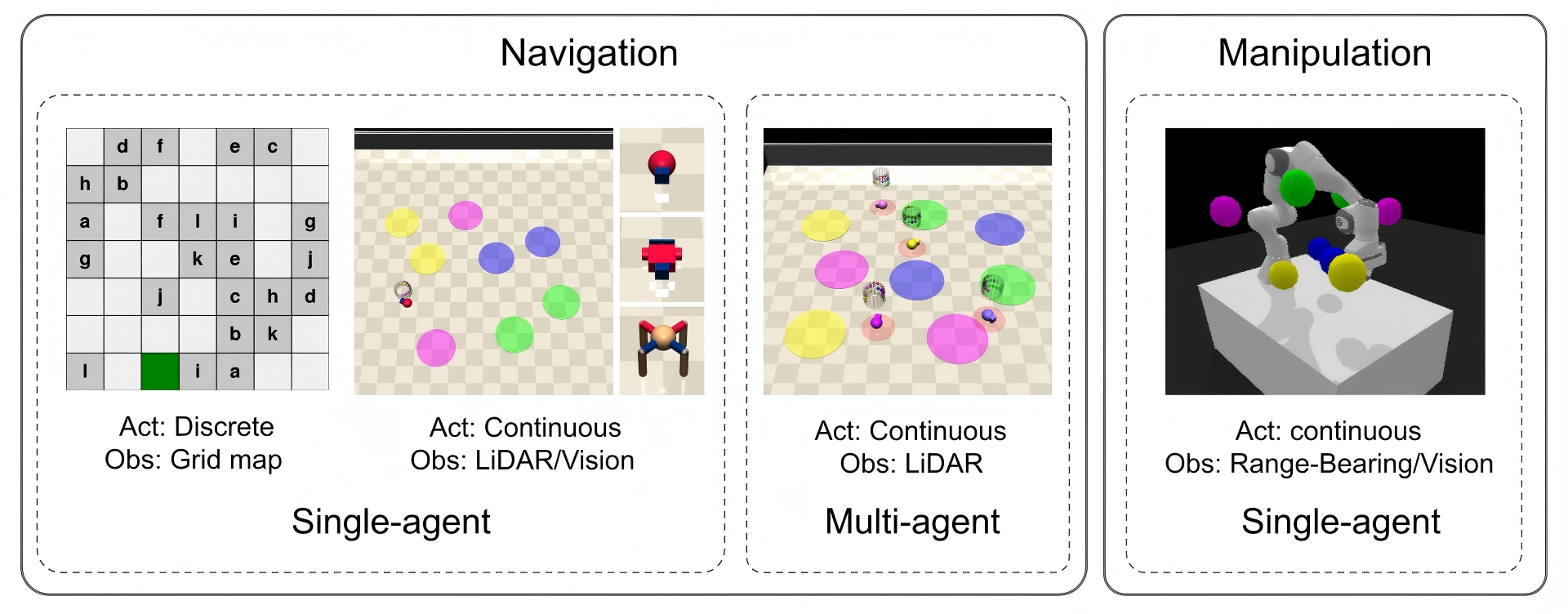}
    \vspace{-5mm}
    \caption{\textbf{SpecRLBench overview.} SpecRLBench spans navigation and manipulation domains with single-agent and multi-agent settings, covering both discrete and continuous action spaces and multiple observation modalities.
    This breadth supports tasks of varying difficulty and enables evaluation of current methods under diverse settings.}
    \label{fig:overview}
\end{figure}

\noindent 
As illustrated in Figure~\ref{fig:overview}, the benchmark includes navigation and manipulation tasks in both single-agent and multi-agent settings.
The environments cover static and dynamic scenarios with multiple robot dynamics and observation modalities.
Together, these design choices enable the construction of task variants with different difficulty levels, facilitating evaluation of performance in terms of generalization to unseen specifications and scalability to environment variations.

\vspace{-2mm}
\vspace{-8mm}

\subsection{Navigation Tasks}
\label{sec:nav}
\textbf{Environments.} The navigation tasks include two environments: \texttt{LetterWorld} and \texttt{ZoneEnv}, adapted from prior work~\citep{vaezipoor2021ltl2action, qiu2023instructing, jackermeier2025deepltl, guo2025one}.
In both environments, the task is to reach target letters or regions while avoiding others, as specified by an LTL formula.
The atomic propositions are by default defined as the letters, e.g., $AP = \{\texttt{a}, \texttt{b}, \cdots\}$, or the colored regions, e.g., $AP = \{ \texttt{green}, \texttt{blue}, \cdots \}$.
Users may also specify custom atomic-proposition sets through environment initialization, enabling evaluation of scalability with respect to the number of propositions.
To promote generality, the spatial locations of letters or regions are randomly sampled at the beginning of each episode during both training and evaluation.
In \texttt{LetterWorld}, the action space is discrete, consisting of \texttt{up}, \texttt{down}, \texttt{left}, and \texttt{right}, and observations are provided as agent-centric grid maps indicating letter locations.
In \texttt{ZoneEnv}, we consider multiple robot embodiments with different dynamics, including \texttt{Point}, a simple planar robot with turning and translation; \texttt{Car}, a wheeled robot with differential-drive control; and \texttt{Ant}, a quadrupedal robot.
The observation space in \texttt{ZoneEnv} supports both state-based inputs, e.g., LiDAR, and pixel-based inputs, e.g., vision, with the observation modality specified by the user at initialization, allowing evaluation across different sensing modalities.

\vspace{2mm}
\noindent \textbf{Variants.} We include several variants of the environments to capture different sources of complexity.
In \texttt{LetterWorld}, partial observability can be enabled by restricting the agent to a limited sensing range instead of providing access to the full grid map.
In \texttt{ZoneEnv}, the regions associated with each atomic proposition are static by default; we additionally consider variants with dynamically moving regions to model more realistic, time-varying settings.
In these dynamic variants, regions may temporarily overlap and later separate, which allows evaluation of policy robustness under changing environment dynamics and non-stationary task conditions.
Users can further specify customized numbers of static and dynamic regions when initializing the environments.
We also construct a multi-agent variant of \texttt{ZoneEnv} as LTL naturally supports the specification of multi-agent behaviors.
In this setting, multiple agents operate simultaneously in a shared environment.
Each agent observes the colored regions and the positions of other agents via LiDAR, allowing the specification to encode coordination, synchronization, and ordering constraints among agents.
For example, each agent can be assigned an individual sequence of target regions, while at certain stages they must coordinate to reach the same region or different regions at the same time, and then resume their own tasks.
This variant enables evaluation of specification-guided RL methods in multi-agent scenarios with shared objectives and inter-agent interactions.

\subsection{Manipulation Tasks}
\label{sec:manip}
\textbf{Environments.} The manipulation environments are adapted from~\citep{gallouedec2021pandagym}.
Similar to the \texttt{ZoneEnv}, different colored regions are placed in the 3D workspace of a 7-DoF robot arm as shown in Figure~\ref{fig:overview}.
The task is to reach or avoid specified regions according to a given LTL formula.
We consider two operation modes with different granularities of specification.
In the first mode, the task specification focuses solely on the grippers of the robotic arm, following prior work~\citep{bagatella2025directed, fan2025imitation}.
In this case, atomic propositions are defined over regions that the grippers should reach or avoid, e.g.,
$AP = \{\texttt{green}, \texttt{blue}, \ldots\}$.
In the second mode, the specification jointly constrains both the robotic arm and the grippers.
Accordingly, atomic propositions distinguish between different robot components and regions, e.g.,
$AP = \{\texttt{grippers\_green}$, $\texttt{arm\_green}$, $\texttt{grippers\_blue}$, $\texttt{arm\_blue}$, $\ldots\}$.
This mode more closely reflects real-world manipulation scenarios in which specifications impose constraints on multiple components of the robot simultaneously.
In both modes, a range-bearing observation is used, providing the distance and direction to the colored regions.
Note that in this manipulation task, we omit explicit grasping and placing behaviors.
While such behaviors are important components of real-world manipulation, the goal of this environment is to evaluate temporal logic requirements rather than low-level manipulation skills of the grippers.
Grasping and placing can be naturally incorporated as extensions, since these behaviors typically build upon reaching actions; for example, the gripper must first reach a target location near an object before grasping and subsequently placing it elsewhere.

\subsection{Environment Interfaces}
\label{sec:env-interfaces}
Our benchmark environments follow a standard Gym-compatible interface~\citep{towers2024gymnasium}, enabling seamless integration with existing RL libraries. 
Each observation is structured into two components: a proposition-dependent part $s_{AP}$, which captures information relevant to atomic propositions, e.g., observations of colored zones, and a proposition-independent part $s_{\neq AP}$, which encodes the agent’s ego-state, e.g., proprioceptive or kinematic features. 
This decomposition provides users with the flexibility to use these components jointly or separately, depending on the requirements of the learning algorithm.
The labeling function is implemented by evaluating the underlying task state. 
In navigation tasks, atomic propositions are determined by checking the agent’s position relative to designated regions, whereas in manipulation tasks, propositions are derived from contact information and interaction states of the robotic arm. At each timestep, the environment returns the set of propositions that are currently satisfied.
To avoid imposing method-specific assumptions, the benchmark does not prescribe a particular reward structure. 
Instead, the default reward is set to zero, allowing users to define rewards or shaping mechanisms appropriate for their learning objectives while relying on the benchmark’s ground-truth proposition assignments and specification satisfaction signals for evaluation.
More details of the environments can be found in Appendix~\ref{sec:app-envs}

\section{Experiments and Analysis}
\label{sec:exps}

\textbf{Baselines.} In this section, we compare representative specification-guided RL methods that target generalization across LTL specifications using our benchmark. 
Specifically, we consider: (1) \textsc{LTL2Action} \citep{vaezipoor2021ltl2action}, which applies LTL progression to track the remaining portion of a specification online and employs a graph neural network to encode its evolving structure; 
(2) \textsc{GCRL-LTL}~\citep{qiu2023instructing}, which uses a heuristic procedure to identify subgoal sequences for satisfying new specifications; 
(3) \textsc{DeepLTL}~\citep{jackermeier2025deepltl}, which learns policies conditioned on embeddings of subgoal sequences; 
(4) \textsc{GenZ-LTL}~\citep{guo2025one}, which conditions policies only on the current subgoal and satisfies specifications incrementally one subgoal at a time; 
and (5) \textsc{RAD-Embeddings}~\citep{yalcinkaya2024compositional}, which pre-trains embeddings over compositions of deterministic finite automata (DFAs) and learns policies conditioned on these automaton-based representations.
We use their official implementation and make modifications to the parameters for fair comparison.
Details can be found in Appendix~\ref{sec:app-baselines}.

\vspace{2mm}
\noindent
\textbf{LTL specifications.} For evaluation, we consider a wide range of LTL specifications, including both finite-horizon and infinite-horizon tasks that cover safety, liveness, and their combinations. 
In both settings, we include: (1) reach-only specifications, where goals are expressed in a nested sequential form of the "eventually" operator; (2) reach-avoid specifications, which combine goal-reaching with safety constraints through nested compositions of negation, "until", and "eventually"; and (3) more complex specifications that mix reach-only and reach-avoid components to capture richer temporal structure. 
In addition, for multi-agent settings, we consider specifications that capture both joint and independent behaviors among agents.
These diverse specifications enable evaluation across varying levels of temporal complexity, safety requirements, and coordination structure.
Some examples of the specifications are listed in Table~\ref{tab:ltl-spec-part}, and the complete list is provided in Appendix~\ref{sec:app-specs}.

\vspace{-3mm}
\begin{table*}[ht]

\caption{Examples of LTL specifications used for evaluation. For specifications in \texttt{Zone} and \texttt{Arm} (grippers-only), $\mathsf{b}$, $\mathsf{g}$, $\mathsf{m}$, and $\mathsf{y}$ denote $\mathsf{blue}$, $\mathsf{green}$, $\mathsf{magenta}$, and $\mathsf{yellow}$, respectively.
For multi-agents in \texttt{Zone}, the number in the atomic proposition denotes the index of the agent.
}
\vspace{-3mm}
\label{tab:ltl-spec-part}
\centering
\renewcommand{\arraystretch}{1.1}
\resizebox{0.95\linewidth}{!}{
\LARGE
\begin{tabular}{cccll}
\Xhline{2.5pt}
                                                                         &                                                                              &                       & \multicolumn{1}{c}{Letter}                                                                                                                                                                                                                                                            & \multicolumn{1}{c}{Zone / Arm (grippers-only)}                                                                                                                                                    \\ \hline
\multirow{5}{*}{\begin{tabular}[c]{@{}c@{}}Single-\\ agent\end{tabular}} & \multirow{2}{*}{\begin{tabular}[c]{@{}c@{}}Finite-\\ horizon\end{tabular}}   & $\Phi_{\text{IND}}$   & $\neg(\mathsf{c} \lor \mathsf{d}) \until ((\neg \mathsf{e} \until \mathsf{l}) \land (\event \mathsf{g}))$                                                                                                                                                                             & $\neg \mathsf{g} \until ((\mathsf{b} \lor \mathsf{m}) \land (\neg \mathsf{g} \until \mathsf{y}))$                                                                                                 \\
                                                                         &                                                                              & $\Phi_{\text{OOD}}$   & $\neg \mathsf{b} \until ((\mathsf{c} \lor \mathsf{d}) \land (\neg \mathsf{e} \until (\mathsf{f} \land \event (\mathsf{g} \land (\neg \mathsf{h} \until (\mathsf{i} \land \event \mathsf{l}))))))$                                                                                     & $\event ((\mathsf{b} \lor \mathsf{g}) \land (\neg \mathsf{y} \until (\mathsf{b} \land (\neg \mathsf{g} \until \mathsf{m})))) \land \event (\mathsf{y} \land (\neg \mathsf{b} \until \mathsf{g}))$ \\ \cline{2-5} 
                                                                         & \multirow{3}{*}{\begin{tabular}[c]{@{}c@{}}Infinite-\\ horizon\end{tabular}} & $\Phi_{\text{rsp}}$   & $(\always \event \mathsf{a}) \land \always (\mathsf{a} \rightarrow (\event (\mathsf{b} \land \event \mathsf{c}) \land (\neg \mathsf{d} \until \mathsf{e}))) \land \always \neg (\mathsf{f} \lor \mathsf{g} \lor \mathsf{h} \lor \mathsf{i})$                                          & $(\always \event \mathsf{b}) \land \always (\mathsf{b} \rightarrow \event \mathsf{g}) \land \always \neg (\mathsf{y} \lor \mathsf{m})$                                                            \\
                                                                         &                                                                              & $\Phi_{\text{rec}}$   & $(\always \event \mathsf{a}) \land (\always \event \mathsf{b}) \land (\always \event \mathsf{c}) \land (\always \event \mathsf{d}) \land (\always \event \mathsf{e}) \land \always \neg (\mathsf{f} \lor \mathsf{g} \lor \mathsf{h} \lor \mathsf{i} \lor \mathsf{j} \lor \mathsf{k})$ & $(\always \event \mathsf{b}) \land (\always \event \mathsf{g}) \land \always \neg (\mathsf{y} \lor \mathsf{m})$                                                                                   \\
                                                                         &                                                                              & $\Phi_{\text{per}}$   & -                                                                                                                                                                                                                                                                                     & $(\event \always \mathsf{y}) \land \always \neg (\mathsf{g} \lor \mathsf{b} \lor \mathsf{m})$                                                                                                     \\ \Xhline{1.5pt}
\multicolumn{3}{l}{}                                                                                                                                                            & \multicolumn{2}{c}{Zone}                                                                                                                                                                                                                                                                                                                                                                                                                                                                  \\ \hline 
\multirow{6}{*}{\begin{tabular}[c]{@{}c@{}}Multi-\\ agent\end{tabular}}  & \multirow{3}{*}{\begin{tabular}[c]{@{}c@{}}Finite-\\ horizon\end{tabular}}   & $\Phi_{\text{indep}}$ & \multicolumn{2}{l}{$(\neg (\mathsf{m\_0} \lor \mathsf{y\_0}) \until (\mathsf{b\_0} \land \event \mathsf{g\_0})) \land (\neg (\mathsf{b\_1} \lor \mathsf{g\_1}) \until (\mathsf{m\_1} \land \event \mathsf{y\_1}))$}                                                                                                                                                                                                                                                                       \\
                                                                         &                                                                              & $\Phi_{\text{coop}}$  & \multicolumn{2}{l}{$\event ((\mathsf{b\_0} \land \mathsf{b\_1}) \land (\neg (\mathsf{m\_0} \lor \mathsf{m\_1}) \until ((\mathsf{y\_0} \land \mathsf{y\_1}) \land \event (\mathsf{g\_0} \land \mathsf{g\_1}))))$}                                                                                                                                                                                                                                                                          \\
                                                                         &                                                                              & $\Phi_{\text{mix}}$   & \multicolumn{2}{l}{$(\neg \mathsf{m\_0} \until \mathsf{y\_0}) \land (\neg \mathsf{b\_1} \until \mathsf{m\_1}) \land \event ((\mathsf{b\_0} \land \mathsf{g\_1}) \land \event (\mathsf{g\_0} \land \mathsf{m\_1}))$}                                                                                                                                                                                                                                                                       \\ \cline{2-5} 
                                                                         & \multirow{3}{*}{\begin{tabular}[c]{@{}c@{}}Infinite-\\ horizon\end{tabular}} & $\Phi_{\text{rsp}}$   & \multicolumn{2}{l}{$(\always \event \mathsf{b\_0}) \land (\always \event \mathsf{g\_1}) \land \always (\mathsf{b\_0} \rightarrow \event \mathsf{y\_1}) \land \always (\mathsf{g\_1} \rightarrow \event \mathsf{m\_0}) \land \always \neg (\mathsf{y\_0} \lor \mathsf{b\_1})$}                                                                                                                                                                                                             \\
                                                                         &                                                                              & $\Phi_{\text{rec}}$   & \multicolumn{2}{l}{$(\always \event (\mathsf{b\_0} \land \mathsf{g\_1})) \land (\always \event (\mathsf{g\_0} \land \mathsf{y\_1})) \land \always \neg (\mathsf{y\_0} \lor \mathsf{m\_1})$}                                                                                                                                                                                                                                                                                               \\
                                                                         &                                                                              & $\Phi_{\text{per}}$   & \multicolumn{2}{l}{$(\event \always (\mathsf{y\_0} \land \mathsf{m\_1})) \land \always \neg (\mathsf{g\_0} \lor \mathsf{b\_0} \lor \mathsf{y\_1} \lor \mathsf{b\_1})$}                                                                                                                                                                                                                                                                                                                    \\ 
                                                                         \Xhline{2.5pt}
\end{tabular}
}
\end{table*}

\noindent
\textbf{Evaluation metrics.}
For finite-horizon and infinite-horizon tasks, the evaluation metrics are:
\begin{itemize}[itemsep=-3pt, topsep=0pt]
    \item (finite) $\eta_s$: success rate, the ratio of trajectories that satisfy the specification.
    \item (finite and infinite) $\eta_v$: violation rate for specifications involving safety constraints.
    \item (finite and infinite) $\eta_o$: others rate, $\eta_o = 1 - \eta_s - \eta_v$. $\eta_s = 0$ for infinite-horizon tasks.
    \item (finite) $\mu$: average number of steps taken to satisfy the specification, computed only over trajectories that satisfy the specification.
    \item (infinite) $\mu_{\text{acc}}$: average number of visits to accepting states, computed only over trajectories that do not violate the specification (involving safety constraints).

\end{itemize}  

\noindent
Since $\mu$ and $\mu_{\text{acc}}$ are conditioned on successful episodes, they are interpreted as a secondary measure of efficiency.
The others rate $\eta_o$ also shows the efficiency of the agent in accomplishing the specified tasks as it captures the fraction of episodes that terminate without either satisfying the specification or violating the safety constraints due to maximum episode length.

\subsection{How do the methods generalize to arbitrary LTL specifications?}
\label{sec:ind-ood-specs}

\textbf{Finite-horizon tasks.} We first evaluate the performance of the baselines on finite-horizon tasks.
We categorize the specifications into in-distribution (IND) specifications $\Phi_{\text{IND}}$, which are sampled from the same distribution used for training, and out-of-distribution (OOD) specifications $\Phi_{\text{OOD}}$, which are unseen during training.
The results are shown in Table~\ref{tab:finite-ind-ood}. 
We can observe a consistent performance degradation when moving from IND to OOD specifications, reflected by reduced success rates and increased violation rates.
The evaluated methods condition policies on different forms of LTL specifications, including their syntax trees, equivalent automata structures, or corresponding subgoal sequences. 
However, the specifications sampled during training are often limited, and performance drops on unseen specifications that induce novel automata structures or longer subgoal sequences.
This highlights the important role of robust specification representations that can capture unseen specifications and support generalization beyond the training distribution.

\vspace{-3mm}
\begin{table*}[h]

\centering
\caption{
Evaluation results of in-distribution $\Phi_{\text{IND}}$ and out-of-distribution $\Phi_{\text{OOD}}$ specifications.
We report the success rate $\eta_s$, violation rate $\eta_v$, and average steps $\mu$ to satisfy the complex specifications listed in Table~\ref{tab:ltl-spec-complete}. 
$\uparrow$: higher is better; $\downarrow$: lower is better.
Each value is averaged over 5 seeds, with 100 trajectories per seed.
}
\vspace{-2mm}
\label{tab:finite-ind-ood}
\LARGE
\begingroup
\setlength{\tabcolsep}{3pt}
\renewcommand{\arraystretch}{1.5}
\resizebox{0.95\linewidth}{!}{

\begin{tabular}{ccccccccccccccccc}
\Xhline{2pt}
\multicolumn{2}{c}{\multirow{2}{*}{}}       & \multicolumn{3}{c}{LTL2Action}                             & \multicolumn{3}{c}{GCRL-LTL}                               & \multicolumn{3}{c}{RAD-embeddings}                         & \multicolumn{3}{c}{DeepLTL}                                & \multicolumn{3}{c}{GenZ-LTL}                               \\
\multicolumn{2}{c}{}                        & $\eta_s \uparrow$ & $\eta_v \downarrow$ & $\mu \downarrow$ & $\eta_s \uparrow$ & $\eta_v \downarrow$ & $\mu \downarrow$ & $\eta_s \uparrow$ & $\eta_v \downarrow$ & $\mu \downarrow$ & $\eta_s \uparrow$ & $\eta_v \downarrow$ & $\mu \downarrow$ & $\eta_s \uparrow$ & $\eta_v \downarrow$ & $\mu \downarrow$ \\ \hline
\multirow{2}{*}{\rotatebox[origin=c]{90}{Letter}} & $\Phi_{\text{IND}}$ & $0.59_{\pm0.09}$  & $0.13_{\pm0.04}$    & $6.46_{\pm0.74}$     & $0.76_{\pm0.09}$  & $0.07_{\pm0.06}$    & $12.23_{\pm2.63}$   & $0.91_{\pm0.06}$  & $0.08_{\pm0.06}$    & $13.63_{\pm3.01}$   & $0.83_{\pm0.07}$  & $0.03_{\pm0.03}$    & $7.32_{\pm1.45}$    & $0.98_{\pm0.03}$  & $0.00_{\pm0.00}$    & $6.82_{\pm1.32}$    \\
                        & $\Phi_{\text{OOD}}$ & $0.01_{\pm0.01}$  & $0.23_{\pm0.14}$    & $18.08_{\pm3.27}$    & $0.66_{\pm0.11}$  & $0.10_{\pm0.10}$    & $25.23_{\pm1.38}$   & $0.79_{\pm0.06}$  & $0.11_{\pm0.05}$    & $30.75_{\pm3.01}$   & $0.70_{\pm0.08}$  & $0.03_{\pm0.03}$    & $14.30_{\pm0.53}$   & $0.94_{\pm0.03}$  & $0.00_{\pm0.00}$    & $13.11_{\pm0.50}$   \\ \hline
\multirow{2}{*}{\rotatebox[origin=c]{90}{Zone}}   & $\Phi_{\text{IND}}$ & $0.58_{\pm0.18}$  & $0.21_{\pm0.13}$    & $405.94_{\pm135.11}$ & $0.85_{\pm0.06}$  & $0.09_{\pm0.06}$    & $377.59_{\pm80.29}$ & $0.92_{\pm0.04}$  & $0.06_{\pm0.05}$    & $366.79_{\pm77.82}$ & $0.84_{\pm0.05}$  & $0.09_{\pm0.05}$    & $273.29_{\pm66.37}$ & $0.98_{\pm0.02}$  & $0.01_{\pm0.01}$    & $252.71_{\pm51.50}$ \\
                        & $\Phi_{\text{OOD}}$ & $0.04_{\pm0.07}$  & $0.40_{\pm0.32}$    & $824.14_{\pm64.46}$  & $0.63_{\pm0.04}$  & $0.11_{\pm0.05}$    & $726.44_{\pm38.83}$ & $0.69_{\pm0.02}$  & $0.15_{\pm0.02}$    & $715.11_{\pm20.63}$ & $0.71_{\pm0.10}$  & $0.13_{\pm0.08}$    & $564.56_{\pm32.40}$ & $0.96_{\pm0.02}$  & $0.01_{\pm0.01}$    & $499.38_{\pm26.37}$ \\ \hline
\multirow{2}{*}{\rotatebox[origin=c]{90}{Arm}}  & $\Phi_{\text{IND}}$ & $0.37_{\pm0.09}$  & $0.13_{\pm0.15}$    & $72.56_{\pm16.75}$   & $0.90_{\pm0.03}$  & $0.10_{\pm0.03}$    & $24.86_{\pm6.18}$   & $0.96_{\pm0.05}$  & $0.01_{\pm0.01}$    & $41.23_{\pm9.04}$   & $0.98_{\pm0.02}$  & $0.01_{\pm0.02}$    & $25.03_{\pm5.95}$   & $1.00_{\pm0.00}$  & $0.00_{\pm0.00}$    & $23.77_{\pm5.27}$   \\
                        & $\Phi_{\text{OOD}}$ & $0.00_{\pm0.00}$  & $0.53_{\pm0.34}$    & -                    & $0.82_{\pm0.07}$  & $0.17_{\pm0.06}$    & $52.80_{\pm0.99}$   & $0.75_{\pm0.09}$  & $0.02_{\pm0.01}$    & $89.82_{\pm9.84}$   & $0.95_{\pm0.03}$  & $0.02_{\pm0.01}$    & $53.28_{\pm3.06}$   & $1.00_{\pm0.01}$  & $0.00_{\pm0.00}$    & $47.81_{\pm1.98}$   \\ \Xhline{2pt}
\end{tabular}

}
\endgroup
\end{table*}

\noindent
\textbf{Infinite-horizon tasks.} We also evaluate the baselines on infinite-horizon tasks. 
Since LTL specifications admit many possible combinations, it is not feasible to evaluate them exhaustively. 
In particular, we consider several representative behavior classes: responsive behaviors $\Phi_{\text{rsp}}$, where visiting one region triggers a requirement to eventually visit another, such as $\always (\alpha_1 \Rightarrow \event \alpha_2)$; recurrence behaviors $\Phi_{\text{rec}}$, which require visiting designated regions infinitely often, such as $\always \event \alpha_1 \land \always \event \alpha_2$, and persistence behaviors $\Phi_{\text{per}}$, which require eventually remaining within a target region, such as $\event \always \alpha_1$.
We also incorporate safety constraints in these specifications, such as $\always \neg \alpha$ to assess whether methods can sustain specification satisfaction over long time horizons.
The evaluated specifications are listed in Table~\ref{tab:ltl-spec-complete}.
Compared with finite-horizon tasks, most baselines exhibit a significant rise in violations, highlighting the increased difficulty of maintaining safety over extended executions.
GCRL-LTL and DeepLTL use a threshold-based heuristic to handle safety constraints, but setting an appropriate threshold can be challenging, as it may vary across specifications.
GenZ-LTL is the only method that explicitly models safety constraints and consequently shows only a slight increase in violation rate.
This emphasizes the importance of effectively modeling safety constraints when learning policies to better satisfy LTL specifications.

\vspace{-4mm}
\begin{table*}[ht]
\centering
\caption{
Evaluation results of different types of infinite-horizon tasks.
We report the violation rate $\eta_v$ and the average number of visits to accepting states $\mu_{\text{acc}}$.
$\uparrow$: higher is better; $\downarrow$: lower is better.
Each value is averaged over 5 seeds, with 100 trajectories per seed.
}
\vspace{-2mm}
\renewcommand{\arraystretch}{1.1}
\resizebox{0.68\linewidth}{!}{

\label{tab:infinite-tasks}
\begin{tabular}{cccccccc}
\Xhline{1.5pt}
\multicolumn{2}{c}{\multirow{2}{*}{}}         & \multicolumn{2}{c}{GCRL-LTL}                      & \multicolumn{2}{c}{DeepLTL}                       & \multicolumn{2}{c}{GenZ-LTL}                      \\
\multicolumn{2}{c}{}                          & $\mu_{\text{acc}} \uparrow$ & $\eta_v \downarrow$ & $\mu_{\text{acc}} \uparrow$ & $\eta_v \downarrow$ & $\mu_{\text{acc}} \uparrow$ & $\eta_v \downarrow$ \\ \hline
\multirow{2}{*}{Letter} & $\Phi_{\text{rsp}}$ & $6.23_{\pm3.02}$            & $0.47_{\pm0.13}$    & $26.00_{\pm5.95}$           & $0.21_{\pm0.04}$    & $60.49_{\pm3.00}$           & $0.01_{\pm0.01}$    \\
                        & $\Phi_{\text{rec}}$ & $2.24_{\pm1.01}$            & $0.56_{\pm0.12}$    & $12.95_{\pm3.44}$           & $0.37_{\pm0.09}$    & $49.54_{\pm2.28}$           & $0.03_{\pm0.03}$    \\ \hline
\multirow{3}{*}{Zone}   
                        & $\Phi_{\text{rsp}}$ & $30.41_{\pm8.00}$           & $0.44_{\pm0.05}$    & $43.87_{\pm12.44}$          & $0.32_{\pm0.10}$    & $55.84_{\pm4.22}$           & $0.05_{\pm0.02}$    \\
                        & $\Phi_{\text{rec}}$ & $30.33_{\pm7.33}$           & $0.46_{\pm0.08}$    & $44.64_{\pm10.34}$          & $0.34_{\pm0.13}$    & $54.91_{\pm4.32}$           & $0.04_{\pm0.02}$    \\ 
                        & $\Phi_{\text{per}}$ & $2047.71_{\pm2618.72}$      & $0.23_{\pm0.08}$    & $7039.35_{\pm1605.49}$      & $0.17_{\pm0.05}$    & $8042.12_{\pm1354.76}$      & $0.04_{\pm0.02}$    \\ \hline
\multirow{3}{*}{Arm}    
                        & $\Phi_{\text{rsp}}$ & $9.59_{\pm0.78}$            & $0.20_{\pm0.02}$    & $9.37_{\pm1.32}$            & $0.06_{\pm0.01}$    & $13.05_{\pm0.58}$           & $0.01_{\pm0.00}$    \\
                        & $\Phi_{\text{rec}}$ & $9.60_{\pm0.45}$            & $0.25_{\pm0.04}$    & $8.40_{\pm1.26}$            & $0.06_{\pm0.02}$    & $12.03_{\pm0.53}$           & $0.00_{\pm0.00}$    \\ 
                        & $\Phi_{\text{per}}$ & $1.88_{\pm0.35}$            & $0.18_{\pm0.02}$    & $0.77_{\pm0.16}$            & $0.03_{\pm0.03}$    & $1.00_{\pm0.29}$            & $0.00_{\pm0.00}$    \\ 
                        \Xhline{1.5pt}
\end{tabular}
}
\end{table*}
\vspace{-2mm}

\vspace{-4mm}
\subsection{How do the methods scale with increasing specification complexity?}
\label{sec:complexity-specs}

\begin{figure}[ht]
    \centering
    \includegraphics[width=1.0\linewidth]{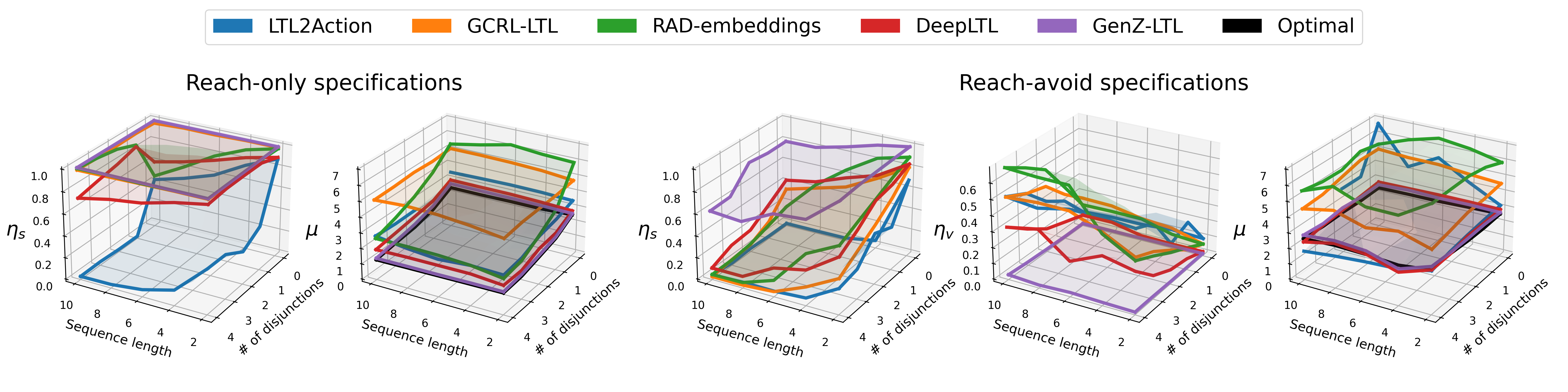}
    \vspace{-10mm}
    \caption{ Evaluation results of reach-only and reach-avoid specifications on \texttt{Letter} with varying sequence length and number of disjunctions of the target to reach/avoid, showing success rate $\eta_s$, violation rate $\eta_v$, and normalized average steps $\mu$. 
    The specifications are shown in Table~\ref{tab:reach-only-reach-avoid-specs}.
    Each value is averaged over 5 seeds, with 100 trajectories per seed.}
    \label{fig:complexity-specs}
\end{figure}

To further evaluate the generalization, we include reach-only and reach-avoid specifications with increasing sequence length.
For reach-only specifications, we additionally vary the number of disjunctions in the reach component, e.g., $\event(\alpha_1 \lor \cdots \lor \alpha_n)$, where the agent can choose among multiple target regions.
For the avoid component in reach-avoid specifications, we similarly vary the number of disjunctions in the safety requirement, e.g., $\neg(\alpha_1 \lor \cdots \lor \alpha_n)\until \alpha_{n+1}$, which requires the agent to avoid multiple regions while progressing toward the target.
We primarily use reach-only specifications to evaluate subgoal selection and efficiency, and reach-avoid specifications to assess the ability to satisfy safety constraints while completing the task.
Although the reach component in reach-avoid specifications could also be varied, doing so would introduce many additional combinations; therefore, we fix it to a single target for simplicity.
The specifications are listed in Table~\ref{tab:reach-only-reach-avoid-specs}, and the results are shown in Figure~\ref{fig:complexity-specs}. 
We report the success rate $\eta_s$, violation rate $\eta_v$, and the average number of steps $\mu$ to satisfy a specification, where $\mu$ is normalized by the sequence length to reflect the efficiency of satisfying each subgoal.
The optimal normalized step values are computed using Dijkstra’s algorithm on the subgoal sequences, leveraging the discrete grid-map observation of \texttt{Letter}, and then averaging the resulting path lengths and normalizing by the sequence length.

\vspace{-2mm}
\vspace{-6mm}

\vspace{2mm}
\noindent
For reach-only specifications, increasing the sequence length requires the agent to satisfy more subgoals in sequence, leading to a decrease in success rate $\eta_s$ for several baselines. 
For example, when the number of disjunctions is fixed to $0$, the success rate consistently drops as the sequence length increases from $2$ to $10$ for LTL2Action, RAD-embeddings, and DeepLTL. 
Success rates also tend to decrease as the number of disjunctions increases, as seen for LTL2Action and DeepLTL, reflecting the OOD challenge where learned representations may not capture unseen formulas well.
From the normalized average steps $\mu$, efficiency often improves with more disjunctions, as agents have additional options for satisfying the specification.
The step metric of GCRL-LTL remains largely unchanged.
It uses a weighted graph search over the automaton, but the learned weights may not always capture the optimal satisfaction path. 
Moreover, while some methods exhibit only small gaps from the optimal normalized steps, others show larger deviations, indicating that longer temporal dependencies and increased disjunctive structure remain challenging and that achieving optimal efficiency is still a key factor in satisfying complex LTL specifications.

\vspace{1.5mm}
\noindent
For reach-avoid specifications, we observe a similar trend in which the success rate decreases as the sequence length increases or as the number of disjunctions grows.
Beyond the OOD issue discussed above, safety constraints play a critical role in this setting.
As the number of disjunctions increases, the specification imposes stronger safety requirements, such as avoiding a larger set of regions while reaching the target. 
Although GenZ-LTL treats safety constraints as hard constraints and achieves the lowest violation rate, its success rate still decreases as specifications become more complex.
In contrast, the remaining methods do not explicitly model safety; while threshold-based heuristics are used to guide actions, the violation rate increases substantially under stricter safety requirements. 
This shows that balancing task objectives, which specify the behaviors to be achieved, and safety constraints, which restrict behaviors to be avoided, as expressed by LTL specifications, remains an important challenge.
In terms of efficiency, all methods continue to exhibit gaps to the optimal policy, indicating room for improvement in efficiently satisfying unseen specifications.


\vspace{-1mm}
\subsection{How do the methods perform in the multi-agent setting?}

\vspace{-5mm}
\begin{table*}[ht]

\centering
\caption{
Evaluation results in multi-agent settings.
For finite-horizon tasks, we report success rate $\eta_s$, violation rate $\eta_v$, and average steps to satisfy the specification $\mu$.
For infinite-horizon tasks, we report violation rate $\eta_v$ and average number of visits to accepting states $\mu_{\text{acc}}$.
$\uparrow$: higher is better; $\downarrow$: lower is better.
}
\label{tab:multi-agent-specs-results}
\vspace{-3mm}
\renewcommand{\arraystretch}{1.2}
\resizebox{0.95\linewidth}{!}{

\begin{tabular}{ccclccclclccclclcccl}
\Xhline{1.5pt}
\multicolumn{2}{c}{\multirow{2}{*}{}}                     & \multicolumn{6}{c}{GCRL-LTL}                                                                                               & \multicolumn{6}{c}{DeepLTL}                                                                                               & \multicolumn{6}{c}{GenZ-LTL}                                                                                              \\
\multicolumn{2}{c}{}                                      & \multicolumn{2}{c}{$\eta_s \uparrow$} & \multicolumn{2}{c}{$\eta_v \downarrow$} & \multicolumn{2}{c}{$\mu \downarrow$}     & \multicolumn{2}{c}{$\eta_s \uparrow$} & \multicolumn{2}{c}{$\eta_v \downarrow$} & \multicolumn{2}{c}{$\mu \downarrow$}    & \multicolumn{2}{c}{$\eta_s \uparrow$} & \multicolumn{2}{c}{$\eta_v \downarrow$} & \multicolumn{2}{c}{$\mu \downarrow$}    \\ \hline
\multirow{3}{*}{Finite-horizon}   & $\Phi_{\text{indep}}$ & \multicolumn{2}{c}{$0.60_{\pm0.07}$}  & \multicolumn{2}{c}{$0.15_{\pm0.04}$}    & \multicolumn{2}{c}{$388.48_{\pm16.03}$}  & \multicolumn{2}{c}{$0.63_{\pm0.06}$}  & \multicolumn{2}{c}{$0.19_{\pm0.08}$}    & \multicolumn{2}{c}{$246.39_{\pm64.40}$} & \multicolumn{2}{c}{$0.80_{\pm0.05}$}  & \multicolumn{2}{c}{$0.06_{\pm0.04}$}    & \multicolumn{2}{c}{$263.27_{\pm56.32}$} \\
                                  & $\Phi_{\text{coop}}$  & \multicolumn{2}{c}{$0.44_{\pm0.11}$}  & \multicolumn{2}{c}{$0.04_{\pm0.03}$}    & \multicolumn{2}{c}{$523.56_{\pm105.66}$} & \multicolumn{2}{c}{$0.51_{\pm0.12}$}  & \multicolumn{2}{c}{$0.02_{\pm0.04}$}    & \multicolumn{2}{c}{$323.68_{\pm72.31}$} & \multicolumn{2}{c}{$0.69_{\pm0.10}$}  & \multicolumn{2}{c}{$0.01_{\pm0.01}$}    & \multicolumn{2}{c}{$325.42_{\pm65.37}$} \\
                                  & $\Phi_{\text{mix}}$   & \multicolumn{2}{c}{$0.63_{\pm0.07}$}  & \multicolumn{2}{c}{$0.09_{\pm0.05}$}    & \multicolumn{2}{c}{$490.37_{\pm92.92}$}  & \multicolumn{2}{c}{$0.57_{\pm0.09}$}  & \multicolumn{2}{c}{$0.17_{\pm0.05}$}    & \multicolumn{2}{c}{$320.97_{\pm30.86}$} & \multicolumn{2}{c}{$0.80_{\pm0.09}$}  & \multicolumn{2}{c}{$0.05_{\pm0.05}$}    & \multicolumn{2}{c}{$324.23_{\pm25.43}$} \\ \hline
                                  &                       & \multicolumn{3}{c}{$\mu_{\text{acc}} \uparrow$}                     & \multicolumn{3}{c}{$\eta_v \downarrow$}                       & \multicolumn{3}{c}{$\mu_{\text{acc}} \uparrow$}                     & \multicolumn{3}{c}{$\eta_v \downarrow$}                      & \multicolumn{3}{c}{$\mu_{\text{acc}} \uparrow$}                     & \multicolumn{3}{c}{$\eta_v \downarrow$}                      \\ \hline
\multirow{3}{*}{Infinite-horizon} & $\Phi_{\text{rsp}}$   & \multicolumn{3}{c}{$1681.99_{\pm2051.82}$}                 & \multicolumn{3}{c}{$0.26_{\pm0.08}$}                          & \multicolumn{3}{c}{$6133.85_{\pm2124.63}$}                 & \multicolumn{3}{c}{$0.17_{\pm0.05}$}                         & \multicolumn{3}{c}{$6338.64_{\pm2482.87}$}                 & \multicolumn{3}{c}{$0.08_{\pm0.03}$}                         \\
                                  & $\Phi_{\text{rec}}$   & \multicolumn{3}{c}{$8.31_{\pm4.16}$}                       & \multicolumn{3}{c}{$0.27_{\pm0.09}$}                          & \multicolumn{3}{c}{$4.82_{\pm2.36}$}                       & \multicolumn{3}{c}{$0.48_{\pm0.21}$}                         & \multicolumn{3}{c}{$14.31_{\pm10.63}$}                     & \multicolumn{3}{c}{$0.07_{\pm0.03}$}                         \\
                                  & $\Phi_{\text{per}}$   & \multicolumn{3}{c}{$4.64_{\pm1.55}$}                       & \multicolumn{3}{c}{$0.25_{\pm0.05}$}                          & \multicolumn{3}{c}{$9.05_{\pm3.90}$}                       & \multicolumn{3}{c}{$0.36_{\pm0.08}$}                         & \multicolumn{3}{c}{$29.31_{\pm7.71}$}                      & \multicolumn{3}{c}{$0.06_{\pm0.03}$}                         \\ 
                                  \Xhline{1.5pt}
\end{tabular}
}
\end{table*}

\vspace{-2mm}
\noindent
In this setting, we consider three types of specifications: independent specifications $\Phi_{\text{indep}}$, which encode separate objectives for individual agents, such as different agents visiting different regions; cooperative specifications $\Phi_{\text{coop}}$, which encode shared objectives among agents, such as requiring all agents to visit the same region simultaneously; and mixed specifications $\Phi_{\text{mix}}$, which combine both independent and cooperative requirements, for example, when one agent follows its own task while all agents are later required to visit the same region.
For infinite-horizon tasks, we evaluate mixed specifications that involve the responsive $\Phi_{\text{rsp}}$, recurrence $\Phi_{\text{rec}}$, and persistence $\Phi_{\text{per}}$ behaviors mentioned in Section~\ref{sec:ind-ood-specs}.
The evaluated specifications are listed in Table~\ref{tab:ltl-spec-complete} and the results are shown in Table~\ref{tab:multi-agent-specs-results}.
Note that all the methods are designed for the single-agent setting; we evaluate the policies trained in the single-agent setting and deploy them directly in multi-agent environments, similar to shared policies for homogeneous agents~\citep{terry2020revisiting}, and collision between agents is not considered.
The multi-agent system receives a system-level specification, and it requires the methods themselves to do the task decomposition and allocation for each agent.


%

\vspace{1.5mm}
\noindent
Among the compared methods, we evaluate GCRL-LTL, DeepLTL and GenZ-LTL, as they can be applied to the multi-agent setting without major modifications.
The whole specification is first converted into a corresponding automaton, from which reach-avoid subgoal sequences are extracted.
We select the shortest path in the automaton to execute.
While this choice may not be optimal, finding the optimal path, i.e., the optimal task allocation among agents, requires further investigations.
We then separate the true assignments on a per-agent basis according to agent indices, and a central coordinator monitors the satisfaction of the overall specification.
For LTL2Action and RAD-Embeddings, generating actions for each agent based on embeddings of the full system-level specification is inefficient, and it is unclear how to decompose a system-level specification into agent-level specifications.
It is also important to note that there exist specifications that may require communication between agents, which the current extended methods cannot satisfy. 
For example, some tasks require all agents to reach their targets at exactly the same time, or require one agent to delay moving to its target until another agent has completed part of its task. 
One agent needs to know the status of other agents to make decisions.
Handling such specifications would require communication mechanisms or coordination-aware policy design. 
A recent work, ACC-MARL~\citep{yalcinkaya2025automata}, targets cooperative tasks by learning a single-agent policy conditioned on embeddings of each agent’s DFA. 
We will evaluate this method when its code becomes available.

\vspace{1.5mm}
\noindent
\textbf{Further experimental results.}
We also evaluate how the baselines perform under increasing environment complexity, including different robot dynamics and control complexity, observation modalities with full or partial observability, and dynamic environments to assess their robustness to environment variations.
Due to page limits, these results are provided in Appendix~\ref{sec:app-complexity-envs}.

\vspace{-1mm}
\section{Conclusion and Directions for Future Work}
\label{sec:conclusion}

In this work, we introduced SpecRLBench, a new benchmark aimed at advancing research on the generalization capabilities of specification-guided RL methods. 
While we have attempted to cover a broad range of factors related to generalization, several limitations remain. 
First, the set of specifications considered in this benchmark is still limited, and there may exist more diverse and challenging specifications under which current methods could fail.
Second, the environments are relatively abstract, e.g., colored zones representing goals, and incorporating environments that are closer to real-world applications would provide a more realistic assessment of practical performance. 
In addition, many challenges in specification-guided RL remain underexplored, such as maintaining high satisfaction rates and strong efficiency on unseen specifications, achieving generalization of specifications in the multi-agent setting, enabling zero-shot or few-shot adaptation to unseen atomic propositions outside the training distribution, and improving sample efficiency during training.
Addressing these challenges is critical for improving the generalization of specification-guided RL.
We hope that the benchmark, analyses, and resources presented in this work provide a useful reference for evaluating future methods, encourage continued exploration in this direction, and ultimately contribute to the development of learning-based agents capable of general-purpose behaviors that can be deployed in real-world applications.

\acks{This work was supported in part by the U.S. National Science Foundation under grant CCF-2340776.}

\bibliography{neus2025}

\clearpage
\appendix
\section{Experimental Settings}

\subsection{Environments}
\label{sec:app-envs}
The \texttt{Letter} environment is a grid world containing 12 letters corresponding to atomic propositions $\text{AP} = \{\text{a}, \text{b}, \ldots, \text{l}\}$. Each letter appears twice and is randomly placed on the grid. The observation consists of either the entire grid or a partial egocentric view, controlled by a flag that enables or disables partial observability. The agent can move in four directions: up, down, left, and right. The grid is wrapped, meaning that if the agent moves out of bounds, it reappears on the opposite side. The maximum episode length is $T = 75$.

\vspace{2mm}
\noindent
The \texttt{Zone} environment contains colored regions corresponding to atomic propositions $\text{AP} = $ $\{\text{blue}$, $\text{green}$, $\text{magenta}$, $\text{yellow}\}$, along with walls that act as boundaries. We implement three types of robots: \texttt{Point}, \texttt{Car}, and \texttt{Ant}. The action space differs across robots: the \texttt{Point} robot controls rotation and forward/backward movement; the \texttt{Car} uses two independently driven parallel wheels with a free-rolling rear wheel; and the \texttt{Ant} controls the torques applied to its leg joints.
Users can also choose between different observation modalities. The observations can consist of LiDAR measurements of the colored regions along with egocentric state information from onboard sensors, or image-based observations captured by a forward-facing camera mounted on the robot. The maximum episode length is $T = 1000$. Users can also enable moving zones in the \texttt{Zone} environment to introduce dynamic elements for evaluation.

\vspace{2mm}
\noindent
For the \texttt{Arm} environment, we implement range–bearing observations that encode the direction and distance to the colored regions. Depending on the selected mode, the observation can include information from the grippers only, or from both the grippers and the robotic arm.

\vspace{2mm}
\noindent
The environments follow the standard Gym APIs, and example code for using them is provided below. The default reward is set to zero.
The true propositions can be obtained from the \texttt{info} field, and rewards can be defined later based on the algorithm design using this ground-truth labeling.
For multi-agent environments, the \texttt{obs} field is a dictionary that contains the observation for each agent.
Users can construct additional \texttt{Wrappers} around the environments to define rewards or termination conditions based on the current true propositions and the specified reach or avoid subgoals.
For all the baselines evaluated, they use the same reward function: $r = 1$ if the "reach" subgoal is achieved and $r=-1$ if the "avoid" subgoal is achieved.

\begin{python}
import gymnasium as gym
import specbench

# Instantiate an environment
env = gym.make('PointLTL0-v0')
# Set seed for reproducibility
seed = 42
# SpecRLBench follows the standard Gym interface
obs, info = env.reset(seed=seed)
# Sample an action and execute it
action = env.action_space.sample()
obs, reward, terminal, timeout, info = env.step(action)
# Retrieve the currently true propositions
active_propositions = info["propositions"]
\end{python}

\subsection{LTL Specifications}
\label{sec:app-specs}

\begin{table*}[ht]

\caption{LTL specifications used for evaluation. For specifications in \texttt{Zone} and \texttt{Arm} (grippers-only), $\mathsf{b}$, $\mathsf{g}$, $\mathsf{m}$, and $\mathsf{y}$ denote $\mathsf{blue}$, $\mathsf{green}$, $\mathsf{magenta}$, and $\mathsf{yellow}$, respectively.
For multi-agents in \texttt{Zone}, the number in the atomic proposition denotes the index of the agent.
}
\vspace{-3mm}
\label{tab:ltl-spec-complete}
\centering
\renewcommand{\arraystretch}{1.25}
\resizebox{0.95\linewidth}{!}{

\begin{tabular}{cccll}
 \Xhline{1.5pt}
\multicolumn{3}{c}{}                                                                                                                                                                              & \multicolumn{1}{c}{Letter}                                                                                                                                                                                                                                                   & \multicolumn{1}{c}{Zone / Arm (grippers-only)}                                                                                                                                                                  \\ \hline
\multirow{30}{*}{\begin{tabular}[c]{@{}c@{}}Single-\\ agent\end{tabular}} & \multirow{10}{*}{\begin{tabular}[c]{@{}c@{}}Finite-\\ horizon\end{tabular}}  & \multirow{5}{*}{$\Phi_{\text{IND}}$}   & $\neg ( \mathsf{a} \lor \mathsf{b}) \until ( \mathsf{c} \land ( \neg ( \mathsf{d} \lor \mathsf{e}) \until \mathsf{f}) )$                                                                                                                                                     & $\neg ( \mathsf{g} \lor \mathsf{y} ) \until ( \mathsf{m} \land ( \neg \mathsf{g} \until \mathsf{b} ) )$                                                                                                         \\
                                                                          &                                                                              &                                        & $( \event \mathsf{b}) \land ( \neg ( \mathsf{c} \lor \mathsf{d}) \until ( \mathsf{e} \land \event \mathsf{f}) )$                                                                                                                                                             & $( \event \mathsf{y} ) \land ( \neg ( \mathsf{y} \lor \mathsf{g} ) \until ( \mathsf{b} \land \event \mathsf{m} ) )$                                                                                             \\
                                                                          &                                                                              &                                        & $\neg ( \mathsf{c} \lor \mathsf{d}) \until ( ( \neg \mathsf{e} \until \mathsf{l}) \land ( \event \mathsf{g}) )$                                                                                                                                                              & $\neg ( \mathsf{g} \lor \mathsf{y} ) \until ( ( \neg \mathsf{g} \until \mathsf{m} ) \land ( \event \mathsf{b} ) )$                                                                                              \\
                                                                          &                                                                              &                                        & $\neg \mathsf{d} \until ( ( \mathsf{e} \lor \mathsf{k}) \land ( \neg \mathsf{g} \until ( \mathsf{h} \land \event \mathsf{i}) ) )$                                                                                                                                            & $\neg \mathsf{g} \until ( ( \mathsf{b} \lor \mathsf{m} ) \land ( \neg \mathsf{g} \until ( \mathsf{y} \land \event \mathsf{b} ) ) )$                                                                             \\
                                                                          &                                                                              &                                        & $\neg ( \mathsf{e} \lor \mathsf{f}) \until ( \mathsf{g} \land \event ( \mathsf{h} \land ( \neg \mathsf{i} \until \mathsf{j}) ) )$                                                                                                                                            & $\neg ( \mathsf{g} \lor \mathsf{m} ) \until ( \mathsf{b} \land \event ( \mathsf{m} \land ( \neg \mathsf{y} \until \mathsf{g} ) ) )$                                                                             \\ \cline{3-5} 
                                                                          &                                                                              & \multirow{5}{*}{$\Phi_{\text{OOD}}$}   & $( \event \mathsf{a}) \land ( \neg ( \mathsf{b} \lor \mathsf{c}) \until ( \mathsf{d} \land \event ( \mathsf{e} \land ( \neg \mathsf{f} \until ( \mathsf{g} \land \event \mathsf{h}) ) ) ) )$                                                                                 & $( \event \mathsf{y} ) \land ( \neg ( \mathsf{y} \lor \mathsf{g} ) \until ( \mathsf{b} \land \event ( \mathsf{m} \land ( \neg \mathsf{y} \until ( \mathsf{g} \land \event \mathsf{b} ) ) ) ) )$                 \\
                                                                          &                                                                              &                                        & $( \neg \mathsf{a} \until \mathsf{b}) \land ( \neg ( \mathsf{c} \lor \mathsf{d}) \until ( \mathsf{e} \land \event ( \mathsf{f} \land ( \neg \mathsf{g} \until ( \mathsf{h} \land \event \mathsf{i}) ) ) ) )$                                                                 & $( \neg \mathsf{g} \until \mathsf{b} ) \land ( \neg ( \mathsf{b} \lor \mathsf{y} ) \until ( \mathsf{m} \land \event ( \mathsf{g} \land ( \neg \mathsf{y} \until ( \mathsf{b} \land \event \mathsf{m} ) ) ) ) )$ \\
                                                                          &                                                                              &                                        & $\neg \mathsf{b} \until ( ( \mathsf{c} \lor \mathsf{d}) \land ( \neg \mathsf{e} \until ( \mathsf{f} \land \event ( \mathsf{g} \land ( \neg \mathsf{h} \until ( \mathsf{i} \land \event \mathsf{l}) ) ) ) ) )$                                                                & $\neg \mathsf{g} \until ( ( \mathsf{b} \lor \mathsf{m} ) \land ( \neg \mathsf{g} \until ( \mathsf{y} \land \event ( \mathsf{b} \land ( \neg \mathsf{y} \until ( \mathsf{m} \land \event \mathsf{b} ) ) ) ) ) )$ \\
                                                                          &                                                                              &                                        & $\neg ( \mathsf{c} \lor \mathsf{d}) \until ( \mathsf{e} \land ( \neg \mathsf{f} \until ( \mathsf{g} \land \event ( \mathsf{h} \land ( \neg \mathsf{i} \until ( \mathsf{j} \land \event \mathsf{k}) ) ) ) ) )$                                                                & $\neg ( \mathsf{y} \lor \mathsf{m} ) \until ( \mathsf{b} \land ( \neg \mathsf{y} \until ( \mathsf{g} \land \event ( \mathsf{m} \land ( \neg \mathsf{g} \until ( \mathsf{y} \land \event \mathsf{b} ) ) ) ) ) )$ \\
                                                                          &                                                                              &                                        & $\neg \mathsf{d} \until ( ( \mathsf{e} \lor \mathsf{f}) \land ( \neg \mathsf{g} \until ( \mathsf{h} \land \event ( \mathsf{i} \land ( \neg \mathsf{j} \until ( \mathsf{k} \land \event \mathsf{l}) ) ) ) ) )$                                                                & $\neg \mathsf{m} \until ( ( \mathsf{y} \lor \mathsf{g} ) \land ( \neg \mathsf{m} \until ( \mathsf{b} \land \event ( \mathsf{m} \land ( \neg \mathsf{b} \until ( \mathsf{g} \land \event \mathsf{y} ) ) ) ) ) )$ \\ \cline{2-5} 
                                                                          & \multirow{9}{*}{\begin{tabular}[c]{@{}c@{}}Infinite-\\ horizon\end{tabular}} & \multirow{3}{*}{$\Phi_{\text{rsp}}$}   & $( \always \event \mathsf{a}) \land \always ( \mathsf{a}-> ( \event ( \mathsf{b} \land \event \mathsf{c}) \land ( \neg \mathsf{d} \until \mathsf{e}) ) ) \land \always \neg ( \mathsf{f} \lor \mathsf{g} \lor \mathsf{h} \lor \mathsf{i})$                                   & $( \always \event \mathsf{b} ) \land \always ( \mathsf{b} -> \event \mathsf{g} ) \land \always \neg ( \mathsf{y} \lor \mathsf{m} )$                                                                             \\
                                                                          &                                                                              &                                        & $( \always \event \mathsf{f}) \land \always ( \mathsf{f}-> ( \event ( \mathsf{e} \land \event \mathsf{d}) \land ( \neg \mathsf{c} \until \mathsf{b}) ) ) \land \always \neg ( \mathsf{a} \lor \mathsf{g} \lor \mathsf{h} \lor \mathsf{i})$                                   & $( \always \event \mathsf{g} ) \land \always ( \mathsf{g} -> \event \mathsf{y} ) \land \always \neg ( \mathsf{b} \lor \mathsf{m} )$                                                                             \\
                                                                          &                                                                              &                                        & $( \always \event \mathsf{i}) \land \always ( \mathsf{i}-> ( \event ( \mathsf{h} \land \event \mathsf{g}) \land ( \neg \mathsf{b} \until \mathsf{a}) ) ) \land \always \neg ( \mathsf{c} \lor \mathsf{d} \lor \mathsf{e} \lor \mathsf{f})$                                   & $( \always \event \mathsf{m} ) \land \always ( \mathsf{m} -> \event \mathsf{y} ) \land \always \neg ( \mathsf{g} \lor \mathsf{b} )$                                                                             \\ \cline{3-5} 
                                                                          &                                                                              & \multirow{3}{*}{$\Phi_{\text{rec}}$}   & $\always \event \mathsf{a} \land \always \event \mathsf{b} \land \always \event \mathsf{c} \land \always \event \mathsf{d} \land \always \event \mathsf{e} \land \always \neg ( \mathsf{f} \lor \mathsf{g} \lor \mathsf{h} \lor \mathsf{i} \lor \mathsf{j} \lor \mathsf{k})$ & $\always \event \mathsf{b} \land \always \event \mathsf{g} \land \always \neg ( \mathsf{y} \lor \mathsf{m} )$                                                                                                   \\
                                                                          &                                                                              &                                        & $\always \event \mathsf{f} \land \always \event \mathsf{g} \land \always \event \mathsf{h} \land \always \event \mathsf{i} \land \always \event \mathsf{j} \land \always \neg ( \mathsf{a} \lor \mathsf{b} \lor \mathsf{c} \lor \mathsf{d} \lor \mathsf{e} \lor \mathsf{k})$ & $\always \event \mathsf{g} \land \always \event \mathsf{y} \land \always \neg ( \mathsf{b} \lor \mathsf{m} )$                                                                                                   \\
                                                                          &                                                                              &                                        & $\always \event \mathsf{k} \land \always \event \mathsf{a} \land \always \event \mathsf{b} \land \always \event \mathsf{c} \land \always \event \mathsf{d} \land \always \neg ( \mathsf{e} \lor \mathsf{f} \lor \mathsf{g} \lor \mathsf{h} \lor \mathsf{i} \lor \mathsf{j})$ & $\always \event \mathsf{m} \land \always \event \mathsf{y} \land \always \neg ( \mathsf{g} \lor \mathsf{b} )$                                                                                                   \\ \cline{3-5} 
                                                                          &                                                                              & \multirow{3}{*}{$\Phi_{\text{per}}$}   & \multirow{3}{*}{-}                                                                                                                                                                                                                                                           & $\event \always \mathsf{y} \land \always \neg ( \mathsf{g} \lor \mathsf{b} \lor \mathsf{m} )$                                                                                                                   \\
                                                                          &                                                                              &                                        &                                                                                                                                                                                                                                                                              & $\event \always \mathsf{g} \land \always \neg ( \mathsf{y} \lor \mathsf{b} \lor \mathsf{m} )$                                                                                                                   \\
                                                                          &                                                                              &                                        &                                                                                                                                                                                                                                                                              & $\event \always \mathsf{b} \land \always \neg ( \mathsf{y} \lor \mathsf{g} \lor \mathsf{m} )$                                                                                                                   \\ \cline{2-5} 
                                                                          & \multicolumn{2}{l}{}                                                                                                  & \multicolumn{2}{c}{Arm (grippers-arm)}                                                                                                                                                                                                                                                                                                                                                                                                                                                         \\ \cline{2-5} 
                                                                          & \multirow{10}{*}{\begin{tabular}[c]{@{}c@{}}Finite-\\ horizon\end{tabular}}  & \multirow{5}{*}{$\Phi_{\text{IND}}$}   & \multicolumn{2}{l}{$\neg ( \mathsf{a_g} \lor \mathsf{a_y} ) \until ( \mathsf{g_m} \land ( \neg \mathsf{a_g} \until \mathsf{g_b} ) )$}                                                                                                                                                                                                                                                                                                                                                          \\
                                                                          &                                                                              &                                        & \multicolumn{2}{l}{$( \event \mathsf{g_y} ) \land ( \neg ( \mathsf{a_y} \lor \mathsf{a_g} ) \until ( \mathsf{g_b} \land \event \mathsf{g_m} ) )$}                                                                                                                                                                                                                                                                                                                                              \\
                                                                          &                                                                              &                                        & \multicolumn{2}{l}{$\neg ( \mathsf{a_g} \lor \mathsf{a_y} ) \until ( ( \neg \mathsf{a_g} \until \mathsf{g_m} ) \land ( \event \mathsf{g_b} ) )$}                                                                                                                                                                                                                                                                                                                                               \\
                                                                          &                                                                              &                                        & \multicolumn{2}{l}{$\neg \mathsf{a_g} \until ( ( \mathsf{g_b} \lor \mathsf{g_m} ) \land ( \neg \mathsf{a_g} \until ( \mathsf{g_y} \land \event \mathsf{g_b} ) ) )$}                                                                                                                                                                                                                                                                                                                            \\
                                                                          &                                                                              &                                        & \multicolumn{2}{l}{$\neg ( \mathsf{a_y} \lor \mathsf{a_m} ) \until ( \mathsf{g_b} \land \event ( \mathsf{g_m} \land ( \neg \mathsf{a_y} \until \mathsf{g_g} ) ) )$}                                                                                                                                                                                                                                                                                                                            \\ \cline{3-5} 
                                                                          &                                                                              & \multirow{5}{*}{$\Phi_{\text{OOD}}$}   & \multicolumn{2}{l}{$( \event \mathsf{g_y} ) \land ( \neg ( \mathsf{a_y} \lor \mathsf{a_g} ) \until ( \mathsf{g_b} \land \event ( \mathsf{g_m} \land ( \neg \mathsf{a_y} \until ( \mathsf{g_g} \land \event \mathsf{g_b} ) ) ) ) )$}                                                                                                                                                                                                                                                            \\
                                                                          &                                                                              &                                        & \multicolumn{2}{l}{$( \neg \mathsf{a_g} \until \mathsf{g_b} ) \land ( \neg ( \mathsf{a_b} \lor \mathsf{a_y} ) \until ( \mathsf{g_m} \land \event ( \mathsf{g_g} \land ( \neg \mathsf{a_y} \until ( \mathsf{g_b} \land \event \mathsf{g_m} ) ) ) ) )$}                                                                                                                                                                                                                                          \\
                                                                          &                                                                              &                                        & \multicolumn{2}{l}{$\neg \mathsf{a_g} \until ( ( \mathsf{g_b} \lor \mathsf{g_m} ) \land ( \neg \mathsf{a_g} \until ( \mathsf{g_y} \land \event ( \mathsf{g_b} \land ( \neg \mathsf{a_y} \until ( \mathsf{g_m} \land \event \mathsf{g_b} ) ) ) ) ) )$}                                                                                                                                                                                                                                          \\
                                                                          &                                                                              &                                        & \multicolumn{2}{l}{$\neg ( \mathsf{a_y} \lor \mathsf{a_m} ) \until ( \mathsf{g_b} \land ( \neg \mathsf{a_y} \until ( \mathsf{g_g} \land \event ( \mathsf{g_m} \land ( \neg \mathsf{a_g} \until ( \mathsf{g_y} \land \event \mathsf{g_b} ) ) ) ) ) )$}                                                                                                                                                                                                                                          \\
                                                                          &                                                                              &                                        & \multicolumn{2}{l}{$\neg \mathsf{a_m} \until ( ( \mathsf{g_y} \lor \mathsf{g_g} ) \land ( \neg \mathsf{a_m} \until ( \mathsf{g_b} \land \event ( \mathsf{g_m} \land ( \neg \mathsf{a_b} \until ( \mathsf{g_g} \land \event \mathsf{g_y} ) ) ) ) ) )$}                                                                                                                                                                                                                                          \\ \hline
\multicolumn{3}{l}{}                                                                                                                                                                              & \multicolumn{2}{c}{Zone}                                                                                                                                                                                                                                                                                                                                                                                                                                                                       \\ \hline
\multirow{18}{*}{\begin{tabular}[c]{@{}c@{}}Multi-\\ agent\end{tabular}}  & \multirow{9}{*}{\begin{tabular}[c]{@{}c@{}}Finite-\\ horizon\end{tabular}}   & \multirow{3}{*}{$\Phi_{\text{indep}}$} & \multicolumn{2}{l}{$( \neg ( \mathsf{m_0} \lor \mathsf{y_0} ) \until ( \mathsf{b_0} \land \event \mathsf{g_0} ) ) \land ( \neg ( \mathsf{b_1} \lor \mathsf{g_1} ) \until ( \mathsf{m_1} \land \event \mathsf{y_1} ) )$}                                                                                                                                                                                                                                                                        \\
                                                                          &                                                                              &                                        & \multicolumn{2}{l}{$( \event \mathsf{b_0} ) \land ( \neg \mathsf{b_0} \until ( \mathsf{g_0} \land \event \mathsf{y_0} ) ) \land ( \event \mathsf{g_1} ) \land ( \neg \mathsf{g_1} \until ( \mathsf{m_1} \land \event \mathsf{b_1} ) )$}                                                                                                                                                                                                                                                        \\
                                                                          &                                                                              &                                        & \multicolumn{2}{l}{$( \neg \mathsf{g_0} \until ( ( \mathsf{b_0} \lor \mathsf{m_0} ) \land ( \neg \mathsf{g_0} \until \mathsf{y_0} ) ) ) \land ( \neg \mathsf{b_1} \until ( ( \mathsf{g_1} \lor \mathsf{y_1} ) \land ( \neg \mathsf{b_1} \until \mathsf{m_1} ) ) )$}                                                                                                                                                                                                                            \\ \cline{3-5} 
                                                                          &                                                                              & \multirow{3}{*}{$\Phi_{\text{coop}}$}  & \multicolumn{2}{l}{$\neg ( \mathsf{m_0} \lor \mathsf{m_1} ) \until ( ( \mathsf{b_0} \land \mathsf{b_1} ) \land \neg ( \mathsf{y_0} \lor \mathsf{y_1} ) \until ( \mathsf{g_0} \land \mathsf{g_1} ) )$}                                                                                                                                                                                                                                                                                          \\
                                                                          &                                                                              &                                        & \multicolumn{2}{l}{$\event ( ( \mathsf{b_0} \land \mathsf{b_1} ) \land ( \neg ( \mathsf{m_0} \lor \mathsf{m_1} ) \until ( ( \mathsf{y_0} \land \mathsf{y_1} ) \land \event ( \mathsf{g_0} \land \mathsf{g_1} ) ) ) )$}                                                                                                                                                                                                                                                                         \\
                                                                          &                                                                              &                                        & \multicolumn{2}{l}{$\event ( ( \mathsf{b_0} \land \mathsf{b_1} ) \land \event ( ( \mathsf{y_0} \land \mathsf{y_1} ) \land \neg ( \mathsf{m_0} \lor \mathsf{m_1} ) \until ( \mathsf{g_0} \land \mathsf{g_1} ) ) )$}                                                                                                                                                                                                                                                                             \\ \cline{3-5} 
                                                                          &                                                                              & \multirow{3}{*}{$\Phi_{\text{mix}}$}   & \multicolumn{2}{l}{$( \neg \mathsf{m_0} \until \mathsf{y_0} ) \land ( \neg \mathsf{b_1} \until \mathsf{m_1} ) \land \event ( ( \mathsf{b_0} \land \mathsf{g_1} ) \land \event ( \mathsf{g_0} \land \mathsf{m_1} ) )$}                                                                                                                                                                                                                                                                          \\
                                                                          &                                                                              &                                        & \multicolumn{2}{l}{$( \neg \mathsf{y_0} \until ( \mathsf{b_0} \land \event \mathsf{g_0} ) ) \land ( \neg \mathsf{b_1} \until \mathsf{g_1} ) \land ( \neg ( \mathsf{y_0} \lor \mathsf{y_1} ) \until ( \mathsf{b_0} \land \mathsf{b_1} ) )$}                                                                                                                                                                                                                                                     \\
                                                                          &                                                                              &                                        & \multicolumn{2}{l}{$\neg \mathsf{m_0} \until ( \mathsf{b_0} \land \event ( \mathsf{m_1} \land \event ( \mathsf{b_0} \land \mathsf{b_1} ) ) ) \land \neg \mathsf{y_1} \until ( \mathsf{g_1} \land \event ( \mathsf{y_0} \land \event ( \mathsf{b_0} \land \mathsf{b_1} ) ) )$}                                                                                                                                                                                                                  \\ \cline{2-5} 
                                                                          & \multirow{9}{*}{\begin{tabular}[c]{@{}c@{}}Infinite-\\ horizon\end{tabular}} & \multirow{3}{*}{$\Phi_{\text{rsp}}$}   & \multicolumn{2}{l}{$( \always \event \mathsf{b_0} ) \land ( \always \event \mathsf{g_1} ) \land \always ( \mathsf{b_0} -> \event \mathsf{y_1} ) \land \always ( \mathsf{g_1} -> \event \mathsf{m_0} ) \land \always \neg ( \mathsf{y_0} \lor \mathsf{b_1} )$}                                                                                                                                                                                                                                  \\
                                                                          &                                                                              &                                        & \multicolumn{2}{l}{$( \always \event \mathsf{g_0} ) \land ( \always \event \mathsf{m_1} ) \land \always ( \mathsf{g_0} -> \event \mathsf{y_1} ) \land \always ( \mathsf{m_1} -> \event \mathsf{b_0} ) \land \always \neg ( \mathsf{b_1} \lor \mathsf{m_0} )$}                                                                                                                                                                                                                                  \\
                                                                          &                                                                              &                                        & \multicolumn{2}{l}{$( \always \event \mathsf{m_0} ) \land ( \always \event \mathsf{y_1} ) \land \always ( \mathsf{m_0} -> \event \mathsf{g_1} ) \land \always ( \mathsf{y_1} -> \mathsf{g_0} ) \land \always \neg ( \mathsf{m_1} \lor \mathsf{b_0} )$}                                                                                                                                                                                                                                         \\ \cline{3-5} 
                                                                          &                                                                              & \multirow{3}{*}{$\Phi_{\text{rec}}$}   & \multicolumn{2}{l}{$\always \event ( \mathsf{b_0} \land \mathsf{g_1} ) \land \always \event ( \mathsf{g_0} \land \mathsf{y_1} ) \land \always \neg ( \mathsf{y_0} \lor \mathsf{m_1} )$}                                                                                                                                                                                                                                                                                                        \\
                                                                          &                                                                              &                                        & \multicolumn{2}{l}{$\always \event ( \mathsf{g_0} \land \mathsf{y_1} ) \land \always \event ( \mathsf{y_0} \land \mathsf{b_1} ) \land \always \neg ( \mathsf{b_0} \lor \mathsf{m_1} )$}                                                                                                                                                                                                                                                                                                        \\
                                                                          &                                                                              &                                        & \multicolumn{2}{l}{$\always \event ( \mathsf{m_0} \land \mathsf{b_1} ) \land \always \event ( \mathsf{b_0} \land \mathsf{y_1} ) \land \always \neg ( \mathsf{g_0} \lor \mathsf{g_1} )$}                                                                                                                                                                                                                                                                                                        \\ \cline{3-5} 
                                                                          &                                                                              & \multirow{3}{*}{$\Phi_{\text{per}}$}   & \multicolumn{2}{l}{$\event \always ( \mathsf{y_0} \land \mathsf{m_1} ) \land \always \neg ( \mathsf{g_0} \lor \mathsf{b_0} \lor \mathsf{y_1} \lor \mathsf{b_1} )$}                                                                                                                                                                                                                                                                                                                             \\
                                                                          &                                                                              &                                        & \multicolumn{2}{l}{$\event \always ( \mathsf{g_0} \land \mathsf{b_1} ) \land \always \neg ( \mathsf{y_0} \lor \mathsf{b_0} \lor \mathsf{m_1} \lor \mathsf{y_1} )$}                                                                                                                                                                                                                                                                                                                             \\
                                                                          &                                                                              &                                        & \multicolumn{2}{l}{$\event \always ( \mathsf{b_0} \land \mathsf{y_1} ) \land \always \neg ( \mathsf{y_0} \lor \mathsf{g_0} \lor \mathsf{m_1} \lor \mathsf{b_1} )$}                                                                                                                                                                                                                                                                                                                             \\ 
                                                                          \Xhline{1.5pt}
\end{tabular}
}
\end{table*}

\noindent
We evaluate the generalization ability of the baselines on a range of specifications, including both finite-horizon and infinite-horizon tasks, as summarized in Table~\ref{tab:ltl-spec-complete} and Table~\ref{tab:reach-only-reach-avoid-specs}.
In the single-agent setting, we evaluate performance on in-distribution $\Phi_{\text{IND}}$ and out-of-distribution $\Phi_{\text{OOD}}$ specifications, as well as infinite-horizon tasks with response $\Phi_{\text{rsp}}$, recurrence $\Phi_{\text{rec}}$, and persistence $\Phi_{\text{per}}$ behaviors. 
We further study generalization in the \texttt{Letter} environment by varying specification complexity through different sequence lengths and numbers of disjunctions.

\vspace{2mm}
\noindent
In the multi-agent setting, for finite-horizon tasks, we evaluate specifications that encode independent $\Phi_{\text{indep}}$, cooperative $\Phi_{\text{coop}}$, and mixed $\Phi_{\text{mix}}$ behaviors that combine both independent and cooperative requirements. 
For infinite-horizon tasks, we focus on mixed specifications that incorporate response, recurrence, and persistence behaviors.

\vspace{-2mm}

\begin{table*}[ht]

\centering
\caption{Reach-only and reach-avoid specification with varying complexity evaluated in \texttt{Letter}. $n_{\text{seq}}$ denotes the sequence length and $n_{\text{disj}}$ denotes the number of disjunctions.
We provide some example specifications with $n_{\text{seq}} = [2, 4]$ and $n_{\text{disj}} = [0, 1]$.
}
\label{tab:reach-only-reach-avoid-specs}
\vspace{-3mm}
\renewcommand{\arraystretch}{1.1}
\resizebox{0.9\linewidth}{!}{

\begin{tabular}{ccll}
\Xhline{1.5pt}
\multicolumn{2}{c}{}                                               & \multicolumn{1}{c}{$n_{\text{disj}}=0$}                                                                                                                                             & \multicolumn{1}{c}{$n_{\text{disj}}=1$}                                                                                                                                                                                                                             \\ \hline
\multirow{6}{*}{Reach-only}  & \multirow{3}{*}{$n_{\text{seq}}=2$} & $\event ( \mathsf{a} \land \event \mathsf{l} )$                                                                                                                                     & $\event ( ( \mathsf{a} \lor \mathsf{b} ) \land \event ( \mathsf{k} \lor \mathsf{l} ) )$                                                                                                                                                                             \\
                             &                                     & $\event ( \mathsf{d} \land \event \mathsf{g} )$                                                                                                                                     & $\event ( ( \mathsf{c} \lor \mathsf{d} ) \land \event ( \mathsf{g} \lor \mathsf{h} ) )$                                                                                                                                                                             \\
                             &                                     & $\event ( \mathsf{f} \land \event \mathsf{k} )$                                                                                                                                     & $\event ( ( \mathsf{e} \lor \mathsf{f} ) \land \event ( \mathsf{i} \lor \mathsf{j} ) )$                                                                                                                                                                             \\ \cline{2-4} 
                             & \multirow{3}{*}{$n_{\text{seq}}=4$} & $\event ( \mathsf{a} \land \event ( \mathsf{b} \land \event ( \mathsf{c} \land \event \mathsf{d} ) ) )$                                                                             & $\event ( ( \mathsf{a} \lor \mathsf{b} ) \land \event ( ( \mathsf{c} \lor \mathsf{d} ) \land \event ( ( \mathsf{e} \lor \mathsf{f} ) \land \event ( \mathsf{g} \lor \mathsf{h} ) ) ) )$                                                                             \\
                             &                                     & $\event ( \mathsf{e} \land \event ( \mathsf{f} \land \event ( \mathsf{g} \land \event \mathsf{h} ) ) )$                                                                             & $\event ( ( \mathsf{e} \lor \mathsf{f} ) \land \event ( ( \mathsf{g} \lor \mathsf{h} ) \land \event ( ( \mathsf{i} \lor \mathsf{j} ) \land \event ( \mathsf{k} \lor \mathsf{l} ) ) ) )$                                                                             \\
                             &                                     & $\event ( \mathsf{i} \land \event ( \mathsf{j} \land \event ( \mathsf{k} \land \event \mathsf{l} ) ) )$                                                                             & $\event ( ( \mathsf{i} \lor \mathsf{j} ) \land \event ( ( \mathsf{k} \lor \mathsf{l} ) \land \event ( ( \mathsf{a} \lor \mathsf{b} ) \land \event ( \mathsf{c} \lor \mathsf{d} ) ) ) )$                                                                             \\ \hline
\multirow{6}{*}{Reach-avoid} & \multirow{3}{*}{$n_{\text{seq}}=2$} & $\neg \mathsf{a} \until ( \mathsf{b} \land ( \neg \mathsf{c} \until \mathsf{d} ) )$                                                                                                 & $\neg ( \mathsf{a} \lor \mathsf{b} ) \until ( \mathsf{c} \land ( \neg ( \mathsf{d} \lor \mathsf{e} ) \until \mathsf{f} ) )$                                                                                                                                         \\
                             &                                     & $\neg \mathsf{e} \until ( \mathsf{f} \land ( \neg \mathsf{g} \until \mathsf{h} ) )$                                                                                                 & $\neg ( \mathsf{e} \lor \mathsf{f} ) \until ( \mathsf{g} \land ( \neg ( \mathsf{h} \lor \mathsf{i} ) \until \mathsf{j} ) )$                                                                                                                                         \\
                             &                                     & $\neg \mathsf{i} \until ( \mathsf{j} \land ( \neg \mathsf{k} \until \mathsf{l} ) )$                                                                                                 & $\neg ( \mathsf{i} \lor \mathsf{j} ) \until ( \mathsf{k} \land ( \neg ( \mathsf{l} \lor \mathsf{a} ) \until \mathsf{b} ) )$                                                                                                                                         \\ \cline{2-4} 
                             & \multirow{3}{*}{$n_{\text{seq}}=4$} & $\neg \mathsf{a} \until ( \mathsf{b} \land ( \neg \mathsf{c} \until ( \mathsf{d} \land ( \neg \mathsf{e} \until ( \mathsf{f} \land ( \neg \mathsf{g} \until \mathsf{h} ) ) ) ) ) )$ & $\neg ( \mathsf{a} \lor \mathsf{b} ) \until ( \mathsf{c} \land ( \neg ( \mathsf{d} \lor \mathsf{e} ) \until ( \mathsf{f} \land ( \neg ( \mathsf{g} \lor \mathsf{h} ) \until ( \mathsf{i} \land ( \neg ( \mathsf{j} \lor \mathsf{k} ) \until \mathsf{l} ) ) ) ) ) )$ \\
                             &                                     & $\neg \mathsf{e} \until ( \mathsf{f} \land ( \neg \mathsf{g} \until ( \mathsf{h} \land ( \neg \mathsf{i} \until ( \mathsf{j} \land ( \neg \mathsf{k} \until \mathsf{l} ) ) ) ) ) )$ & $\neg ( \mathsf{e} \lor \mathsf{f} ) \until ( \mathsf{g} \land ( \neg ( \mathsf{h} \lor \mathsf{i} ) \until ( \mathsf{j} \land ( \neg ( \mathsf{k} \lor \mathsf{l} ) \until ( \mathsf{a} \land ( \neg ( \mathsf{b} \lor \mathsf{c} ) \until \mathsf{d} ) ) ) ) ) )$ \\
                             &                                     & $\neg \mathsf{i} \until ( \mathsf{j} \land ( \neg \mathsf{k} \until ( \mathsf{l} \land ( \neg \mathsf{a} \until ( \mathsf{b} \land ( \neg \mathsf{c} \until \mathsf{d} ) ) ) ) ) )$ & $\neg ( \mathsf{i} \lor \mathsf{j} ) \until ( \mathsf{k} \land ( \neg ( \mathsf{l} \lor \mathsf{a} ) \until ( \mathsf{b} \land ( \neg ( \mathsf{c} \lor \mathsf{d} ) \until ( \mathsf{e} \land ( \neg ( \mathsf{f} \lor \mathsf{g} ) \until \mathsf{h} ) ) ) ) ) )$ \\ 
                             \Xhline{1.5pt}
\end{tabular}


}
    
\end{table*}


\vspace{-5mm}
\section{Baselines}
\label{sec:app-baselines}
We use the official codebases of the baselines: LTL2Action\footnote{\url{https://github.com/LTL2Action/LTL2Action}}, GCRL-LTL\footnote{\url{https://github.com/RU-Automated-Reasoning-Group/GCRL-LTL}}, RAD-embeddings\footnote{\url{https://github.com/RAD-Embeddings/neurips24}}, DeepLTL\footnote{\url{https://github.com/mathiasj33/deep-ltl}}, and GenZ-LTL\footnote{\url{https://github.com/BU-DEPEND-Lab/GenZ-LTL}}.
We integrate our environments into the original training and evaluation pipelines of the compared methods. 
For fair comparison, all methods share the same policy and critic architectures. The actor network is a fully connected network with three hidden layers of sizes [64, 64, 64], and the critic network has two hidden layers of sizes [64, 64].
We use the \texttt{Adam} optimizer with a learning rate of $3\times10^{-4}$ and train all methods for 15M environment interactions. The discount factor is set to $\gamma=0.94$ for \texttt{Letter}, $\gamma=0.998$ for \texttt{Zone}, and $\gamma=0.99$ for \texttt{Arm}.
For observation encoders, we use CNN models for grid-map observations in \texttt{Letter} and image-based observations in \texttt{Zone}. 
The CNN architecture for \texttt{Letter} uses channels=[16, 32, 64] with kernel size $(2,2)$. 
For \texttt{Zone}, we use channels=[16, 32, 64, 64, 32] with kernel size $(5,5)$, stride $=2$, and padding $=1$.
For LiDAR observations in \texttt{Zone}, we use a fully connected encoder with two hidden layers of sizes [128, 64]. 
For range-bearing observations in \texttt{Arm}, we use fully connected encoders with two hidden layers of sizes [128, 64] for the grippers-only mode and [256, 64] for the grippers-arm mode.
The embeddings of the environment observation and the embeddings related to specifications are concatenated and used as the input to both the actor and critic networks.

\section{Further Experimental Results}
\label{sec:app-further-exps}

\subsection{How do the methods perform under increasing environment complexity?}
\label{sec:app-complexity-envs}

\begin{figure}[ht]
    \centering
    \includegraphics[width=0.9\linewidth, trim=2cm 0cm 2cm 0cm, clip]{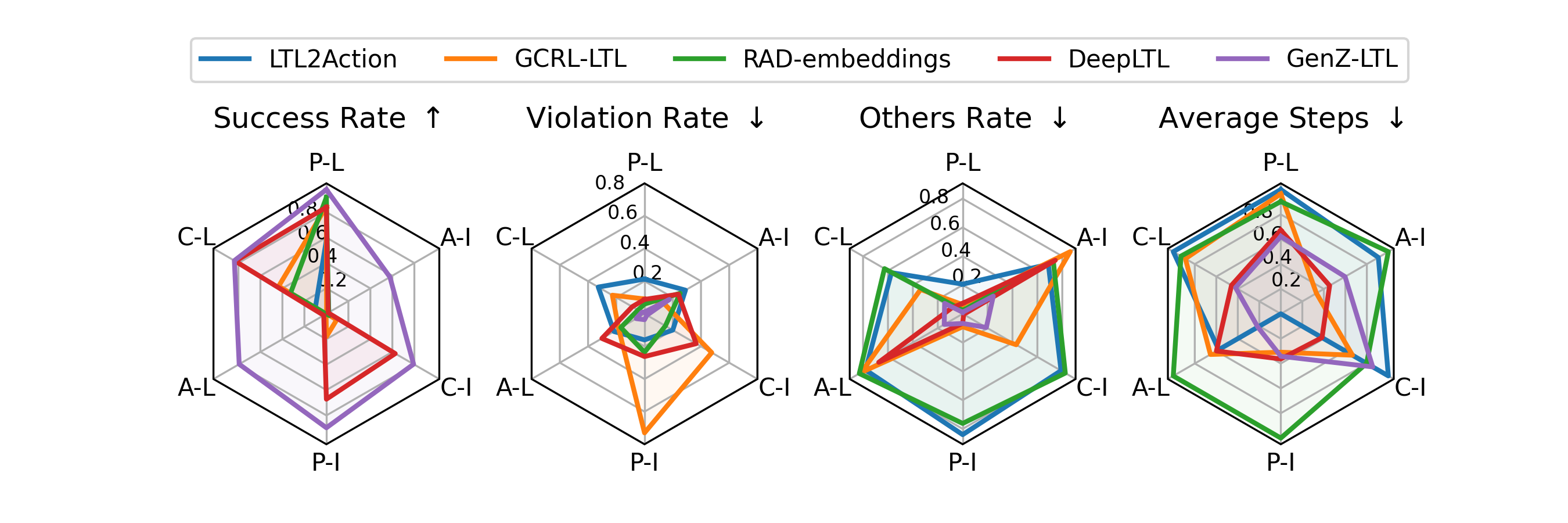}
    \vspace{-6mm}
    \caption{Evaluation results of in-distribution $\Phi_{\text{IND}}$ specifications under different robot dynamics and observation modalities.
    We report the success rate, violation rate, others rate, and normalized average steps to satisfy the specifications. 
    \texttt{P}, \texttt{C}, and \texttt{A} denote the \texttt{Point}, \texttt{Car}, and \texttt{Ant} robots, while L and I denote LiDAR and image-based observations.
    Each value is averaged over 5 seeds, with 100 trajectories per seed.
    }
    \label{fig:dynamics-obs-type}
\end{figure}

\begin{figure}[ht]
    \centering
    \includegraphics[width=0.9\linewidth]{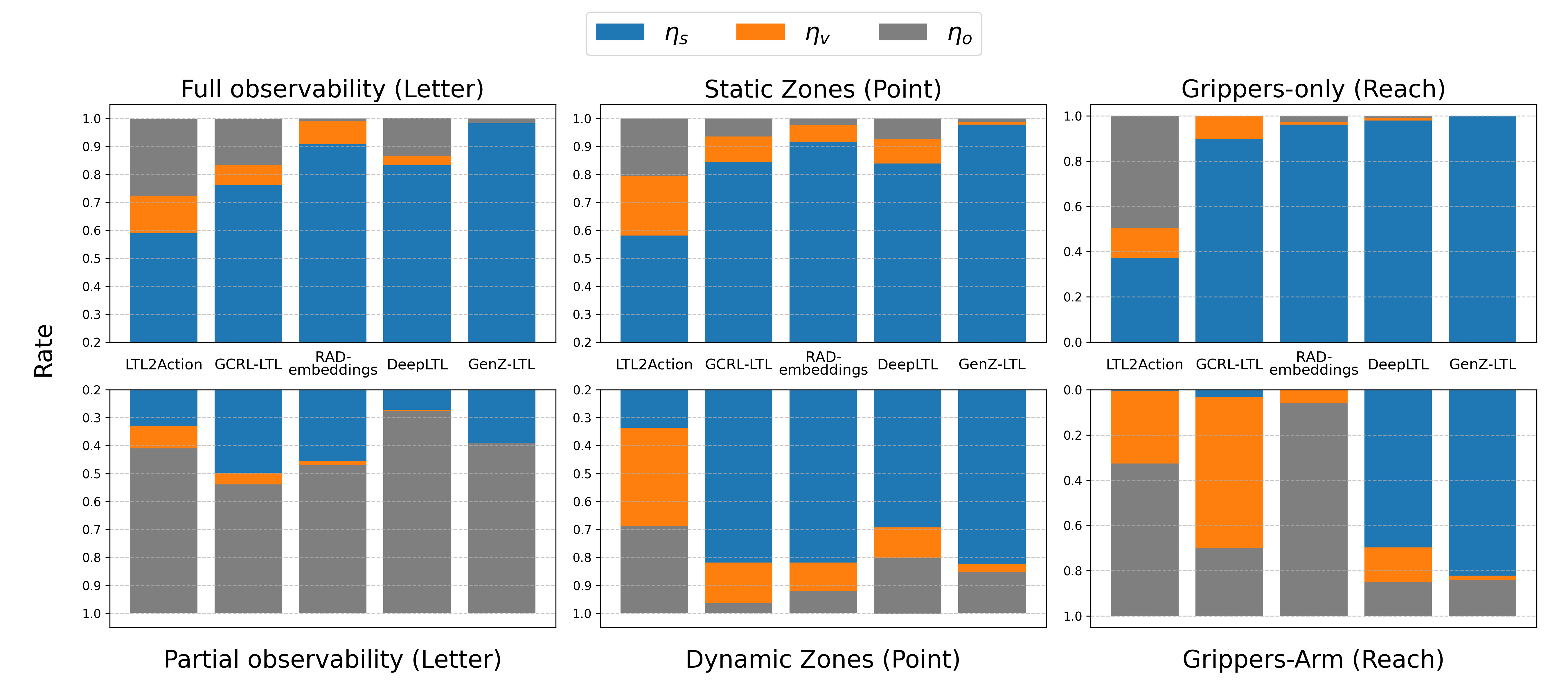}
    \vspace{-2mm}
    \caption{Evaluation results on in-distribution specifications $\Phi_{\text{IND}}$ under different settings: partial observability with limited sensing range (left), dynamic environments with moving zones (middle), and different manipulation modes considering only the grippers or both the grippers and the robotic arm (right). We report the success rate $\eta_s$, violation rate $\eta_v$, and others rate $\eta_o$. 
    Each value is averaged over 5 random seeds, with 100 trajectories per seed.}
    \label{fig:partial-dynamic-grippers}
\end{figure}

\noindent
Last but not least, we evaluate the methods under varying levels of environment complexity, including different robot dynamics and control complexity, observation modalities with full or partial observability, and dynamic environments, to assess their scalability.
For robot dynamics and observation modalities, we use the \texttt{Zone} environment with different robot such as \texttt{Point}, \texttt{Car}, and \texttt{Ant} with LiDAR or image-based observations.
Image observations naturally introduce partial observability, and we also evaluate this aspect in the \texttt{Letter} environment by limiting the range of the grid-map observation.
To vary control complexity, in addition to the differences in robot dynamics, we also include the \texttt{Arm} environment under two configurations: one that considers only the grippers, and another that considers both the grippers and the robotic arm.
The results are shown in Figure~\ref{fig:dynamics-obs-type} and Figure~\ref{fig:partial-dynamic-grippers}. 
For dynamic environments, we include the \texttt{Zone} environment with moving zones.
We can observe several trends. 
First, as the robot dynamics become more complex (from \texttt{Point} to \texttt{Car} to \texttt{Ant}) or the control space becomes more constrained (from grippers-only to grippers–arm), the success rate decreases while the violation rate increases, reflecting the growing difficulty of learning effective control policies. 
Second, as the observation space becomes higher-dimensional (from LiDAR to images) or partially observable (e.g., grid-map observations with a limited sensing range), the success rate further drops and the violation rate increases, as learning the mapping from observations to actions becomes more challenging. 
Third, when the environment becomes dynamic, the task becomes harder due to additional uncertainty, again leading to lower success rates and higher violation rates.
While existing methods can perform well in relatively simple and fully observable settings, their scalability is limited by increased dynamical complexity, constrained control settings, partial observability, and environmental non-stationarity. 

\section{Visualization}
\label{sec:app-vis}

\begin{figure}[ht]
    \centering
    \includegraphics[width=0.85\linewidth]{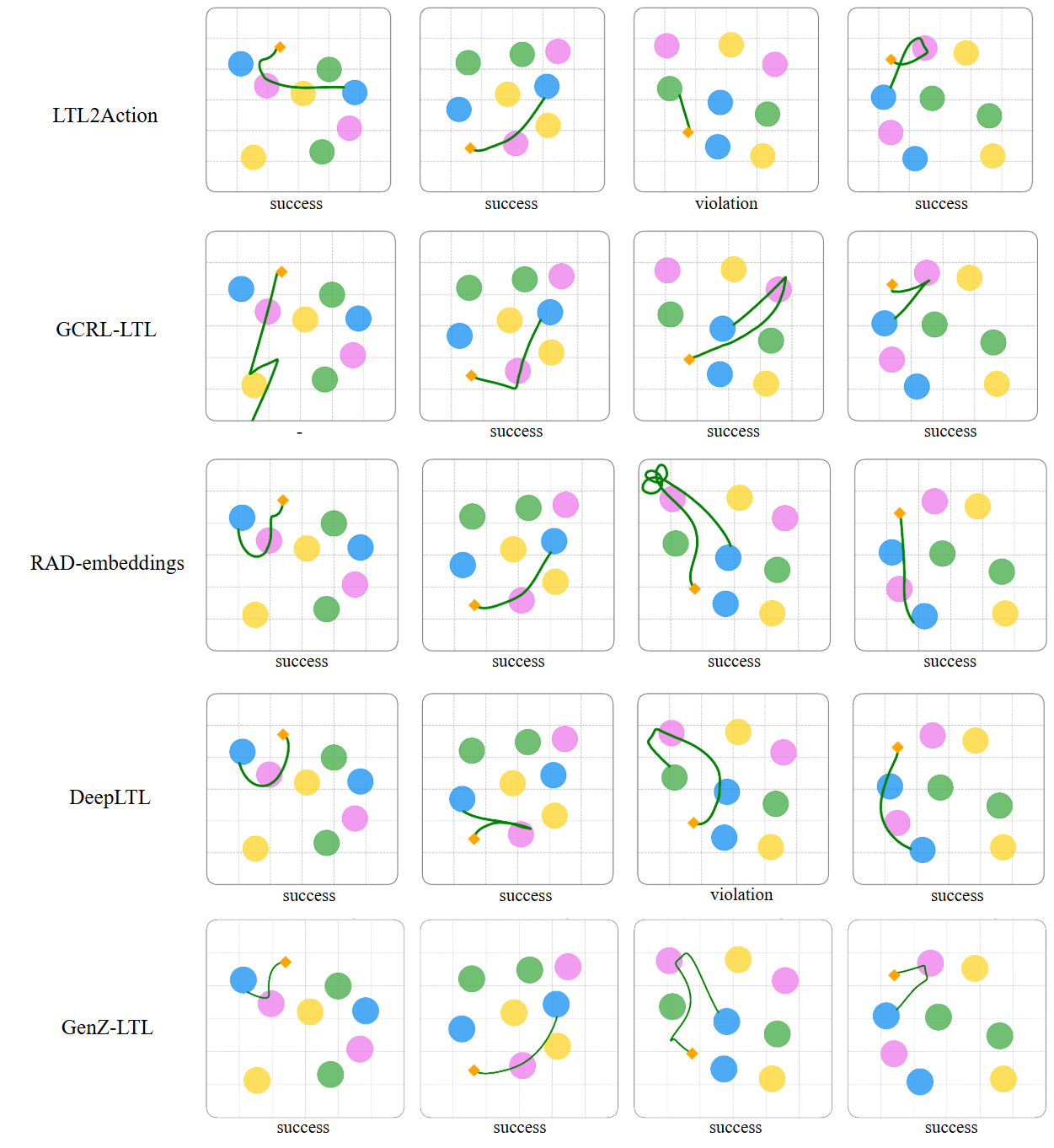}
    \vspace{-3mm}
    \caption{
    Visualization of baseline trajectories for the specification $\neg(\mathsf{g} \lor \mathsf{y}) \until (\mathsf{m} \land (\neg \mathsf{g} \until \mathsf{b}))$. The agent is required to first reach the $\mathsf{magenta}$ region while avoiding $\mathsf{green}$ and $\mathsf{yellow}$ regions, and then reach the $\mathsf{blue}$ region while continuing to avoid $\mathsf{green}$ regions.
    }
    \label{fig:app-vis}
\end{figure}

An illustration of the trajectories of the baselines in the \texttt{Zone} environment is shown in Figure~\ref{fig:app-vis}. The specification is $\neg(\mathsf{g} \lor \mathsf{y}) \until (\mathsf{m} \land (\neg \mathsf{g} \until \mathsf{b}))$, which is listed in Table~\ref{tab:ltl-spec-complete}.
This specification requires the agent to first reach the $\mathsf{magenta}$ region while avoiding $\mathsf{green}$ and $\mathsf{yellow}$ regions, and then proceed to the $\mathsf{blue}$ region while continuing to avoid $\mathsf{green}$.
From the trajectories, we observe that the baselines show different behaviors: some agents violate safety constraints and fail to satisfy the specification, while others follow safe paths but take inefficient paths toward the targets, which indicates that there remains room for improvement in both compliance and efficiency.

\end{document}